%% file: main.tex
\pdfoutput=1

\documentclass[letterpaper,twocolumn,10pt]{article}
\usepackage{usenix2019_v3}

\usepackage{tikz}
\usepackage{amsmath}

\usepackage{filecontents}

\usepackage{epsfig,endnotes,comment}
\usepackage{xcolor}
\usepackage{multirow}
\usepackage{graphicx}
\usepackage[font=small]{caption}
\usepackage[caption=true]{subfig} 
\usepackage{array}
\usepackage[ruled,linesnumbered,boxed]{algorithm2e} 
\usepackage{xspace}
\usepackage{microtype}
\usepackage{url}

\usepackage{cleveref}
\crefname{section}{\S\hspace*{-0.09cm}}{\S\S\hspace*{-0.09cm}}
\Crefname{section}{\S\hspace*{-0.09cm}}{\S\S\hspace*{-0.09cm}}
\crefformat{section}{\S#2#1#3}

\graphicspath{ {fig/} }

\newcommand{\sys}{Stanza\xspace}
\newcommand{\xr}[1]{{\color{red} #1}}
\newcommand{\hx}[1]{{\color{blue} #1}}
\newcommand\Emph{\textbf}

\begin{document}
%-------------------------------------------------------------------------------

%don't want date printed
\date{}

% make title bold and 14 pt font (Latex default is non-bold, 16 pt)
\title{\sys: Layer Separation for Distributed Training in Deep Learning
}

%for single author (just remove % characters)
\author{
{\rm Xiaorui Wu\quad\quad Hong Xu}\\
City University of Hong Kong
% xiaorui.wu@my.cityu.edu.hk
% \and
% {\rm Hong Xu}\\
% City University of Hong Kong
% henry.xu@cityu.edu.hk
\and
{\rm Bo Li}\\
HKUST
% bli@cse.ust.hk
\and
{\rm Yongqiang Xiong}\\
Microsoft Research
% yongqiang.xiong@microsoft.com
% copy the following lines to add more authors
% \and
% {\rm Name}\\
%Name Institution
} % end author

\maketitle

\begin{abstract}
\input{abstract}
\end{abstract}

% \keywords{Deep learning, parameter server, distributed training, layer
% separation}

\input{intro}
\input{background}

\input{motivation}
\input{design}

\input{implement}
\input{evaluation}

\input{numerical}
\input{dis}

\input{related}
\input{conclusion}

%-------------------------------------------------------------------------------
\clearpage
\bibliographystyle{abbrv}
\bibliography{main}

%%%%%%%%%%%%%%%%%%%%%%%%%%%%%%%%%%%%%%%%%%%%%%%%%%%%%%%%%%%%%%%%%%%%%%%%%%%%%%%%
\end{document}

%% file: abstract.tex
%!TEX root = main.tex

The parameter server architecture is prevalently used for distributed deep
learning. Each worker machine in a such system trains the complete model,
which leads to a large amount of network data transfer between workers and
servers. We empirically observe that the data transfer has a major impact on
training time.

We present a new distributed training system called Stanza to tackle this
problem. Stanza exploits the fact that in many models such as convolution
neural networks, most data exchange is attributed to the fully connected
layers, while most computation is carried out in convolutional layers. Thus,
we propose layer separation in distributed training: most nodes of the cluster
train only the convolutional layers, while the rest train the fully connected
layers. Gradients and parameters of the fully connected layers no longer need
to be exchanged across the entire cluster, thereby substantially reducing the
data transfer volume. We implement Stanza on PyTorch and evaluate its
performance on Azure and EC2. Results show that Stanza accelerates training
significantly over current parameter server systems: on EC2 instances with
Tesla V100 GPU and 10Gb bandwidth for example, Stanza is 1.34x--13.9x faster
for common deep learning models.

%% file: intro.tex
%!TEX root = main.tex

\section{Introduction}

Deep learning (DL) has recently achieved prominent success in many
problem domains, particularly image classification 
\cite{krizhevsky2012imagenet,simonyan2014very,szegedy2016rethinking,he2016identity} and speech recognition \cite{swietojanski2014convolutional}. As DL models are made more complicated and data are procured from more sources at a faster pace, distributed training becomes increasingly common in large organizations.
Most distributed DL systems follow the {\em parameter server} architecture first proposed in \cite{smola2010architecture} and later refined in \cite{li2014scaling,ho2013more,xing2015petuum,ahmed2012scalable}. 
A parameter server system has two types of machines: workers and servers. 
Workers perform training on local data shards they are assigned to, and send the gradients of the model parameters to the corresponding servers. 
Servers update the parameters based on aggregated gradients from all workers, and push the latest parameters back to them for the next iteration.

In current parameter server systems workers train the complete DL model, and
exchange the entire set of parameters with the servers in each iteration. This
is termed the {\em data parallelism} paradigm. Given the sheer size of DL
models with hundreds of millions of parameters
\cite{203269,krizhevsky2012imagenet,simonyan2014very,szegedy2016rethinking,he2016identity},
this approach produces an substantial amount of network data transfer. For
example VGG-16 \cite{simonyan2014very}, Inception-V3
\cite{szegedy2016rethinking}, and ResNet \cite{he2016identity} have over
{1100MB, 210MB, and 480MB data, respectively, that each worker has to exchange
with servers in each iteration.} Even with 40G or 100G interconnects the
parameter exchange has a non-negligible impact on training time, especially
when modern GPUs with tens of TFLOPs computation capacity are deployed
\cite{V100}.

% Given the sheer size of models, one intuitive approach to mitigate the communication bottleneck is then to reduce the number of parameters transmitted over the network.  
Most DL models, in particular convolutional neural networks (CNNs), are
composed of two basic building blocks: {convolutional (CONV) layers} and {fully
connected (FC) layers}. 
More interestingly, we find that CONV and FC layers exhibit distinct
characteristics (\cref{sec:layers}). CONV layers are essentially small filters
that extract features through convolution over the training data. 
Thus they usually have a very small number of parameters ($<$10\% of the total),
but need a lot of computation ($>$90\%). On the contrary, FC layers have full
connectivity between layers in order to perform high-level reasoning. 
They generally use most of the parameters in the model, but only require a small
amount of computation. 

Motivated by these observations, we propose to {\em decouple} the training of
CONV and FC layers instead of bundling them as in current parameter server
architecture (\cref{subsec:process}). 
The idea is simple: We use most machines in the cluster as convolutional
workers, or CONV workers, whose job is to train just the CONV layers that
are computationally demanding. A few remaining machines then act as FC
workers that train the FC layers and update their parameters.
The FC layer gradient and parameter exchange, which accounts for most of the
communication cost as discussed, now only happens among a few FC workers,
which reduces the amount of data transfer and training time considerably.  

To demonstrate the feasibility of our idea, we design and implement a new
distributed DL system called \sys (\cref{sec:design}). \sys's design addresses
two new challenges introduced by the separate training of models.

\noindent{\bf Communication Strategy.}
% The first is communication strategy. 
In order for training (with stochastic gradient descent
\cite{bottou2010large,klein2009adaptive,ZWSL10}) to work, CONV workers now
need to send activations of the last CONV layer to FC workers to complete the
forward pass. Similarly FC workers need to push the gradients of the last CONV
layer to CONV workers for backpropagation. Further, without centralized
servers, CONV (resp. FC) workers need to exchange among themselves CONV (resp.
FC) layer gradients.

\sys adopts hybrid communication strategies to minimize the cost of these
communication patterns (\cref{sec:hybrid_Comm}). For CONV-FC communication,
since the number of activations of the last CONV layer is very small compared
to the complete model, \sys simply uses many-to-one and one-to-many
communication here. For the more expensive gradient exchange among CONV
(resp. FC) workers themselves, \sys adopts efficient algorithms for the {\tt
allreduce} operation in the MPI literature \cite{thakur2005optimization} to
fully parallelize the duplex communication among nodes.
\sys also overlaps the CONV worker and FC worker gradient exchange to further reduce the training time. 

\noindent{\bf Node Assignment.}
The second design challenge is how to determine the node assignment in \sys.
That is, how many nodes in the cluster should be used as CONV/FC workers. We
take a principled approach here (\cref{sec:node_assignment}). First we develop
a performance model to characterize \sys's training throughput for a given
node assignment scheme based on the hybrid communication strategies. We then
cast the node assignment problem as an optimization program that aims to
maximize the
training performance given the total number of nodes, which can be solved
offline efficiently.

We implement \sys on PyTorch \cite{paszke2017pytorch}
(\cref{sec:implementation}) and evaluate it with small-scale deployments on
Azure and AWS EC2 (\cref{sec:eval}). Our evaluation uses modern GPUs such as
Nvidia Tesla V100 and widely used CNNs in image classification including
AlexNet \cite{krizhevsky2012imagenet}, VGG-16 and VGG-19
\cite{simonyan2014very}, Inception-V3 \cite{szegedy2016rethinking}, and
ResNet-152 \cite{he2016identity}. We show that \sys delivers significant
speedup in training and saves up to 10x data transfer compared to parameter
server systems. With 10 CONV workers and 1 FC worker for example,
\sys provides 13.9x and 8.2x speedups for AlexNet and VGG-16, respectively,
using a 10G network. Even with 40G or 100G network bandwidth, our numerical
simulation in \cref{sec:simulation} shows that \sys still provides salient
benefits: it achieves 1.55x and 1.72x speedups for AlexNet and VGG-16,
respectively, with 100G bandwidth.

%% file: background.tex
%!TEX root = main.tex

\section{Background }
\label{sec:back}

We start by introducing background on deep learning and the parameter server
architecture.

\subsection{Deep Learning}
\label{sec:dl}
% \begin{figure}[h]
% \includegraphics[scale=0.3]{ENet}
% \centering{}
% \caption{Four layers artificial neural network}
% \label{example network}
% \end{figure}

Deep learning (DL) is a class of machine learning algorithms that uses a large
number of connected layers of different processing functions to handle complex
tasks. The layered structure of DL models composes a large artificial neural
network which resembles the biological structure of the human brain
\cite{lecun2015deep}. The objective of DL is to find a model that minimizes
the difference between the inference result from the model and the ground
truth, which is usually represented by a loss function. This is effectively an
optimization problem and thus many optimization algorithms are used to
iteratively train the DL models.

Particularly, stochastic gradient descent (SGD) is widely used in DL
\cite{bottou2010large,klein2009adaptive,sutskever2013importance,ZWSL10}. The
process consists of two phases with {mini-batch} SGD: forward pass and
backpropagation. In the forward pass of the $t$-th iteration, a batch of input
data $D_t$ is fed to the model, and a loss value $l(x, w_t)$ is computed as
the result for each sample $x$. Here $w_t$ represents model parameters. Then
in backpropagation, the parameters are revised according to the loss so that
the model ``learns'' about the correct answers. Mathematically, the parameter
update rule is:
\begin{equation}
w_{t+1} \longleftarrow w_{t} - \frac \eta n \sum_{x \in D_t} \nabla l(x, w_t), 
\label{SGD}
\end{equation}
where $\eta$ is the learning rate {and $n$ the batch size}. The loss value
is calculated at the output layer of the neural network, and gradients are
generated from the output layer all the way back using the chain rule, so that
parameters at each layer can be updated.

\subsection{Parameter Server Systems}
\label{sec:ps}

DL Training is usually done in a distributed setting with a cluster of
machines in order to cope with the increasingly large datasets and complex
models. The parameter server architecture has emerged as the {\em de facto}
solution for distributed training since its inception
\cite{li2014scaling,xing2015petuum,ahmed2012scalable}, and is implemented in
almost all DL frameworks \cite{abadi2016tensorflow,chen2015mxnet,Paddle}.

\begin{figure}[ht]
\centering
\includegraphics[width=1\linewidth]{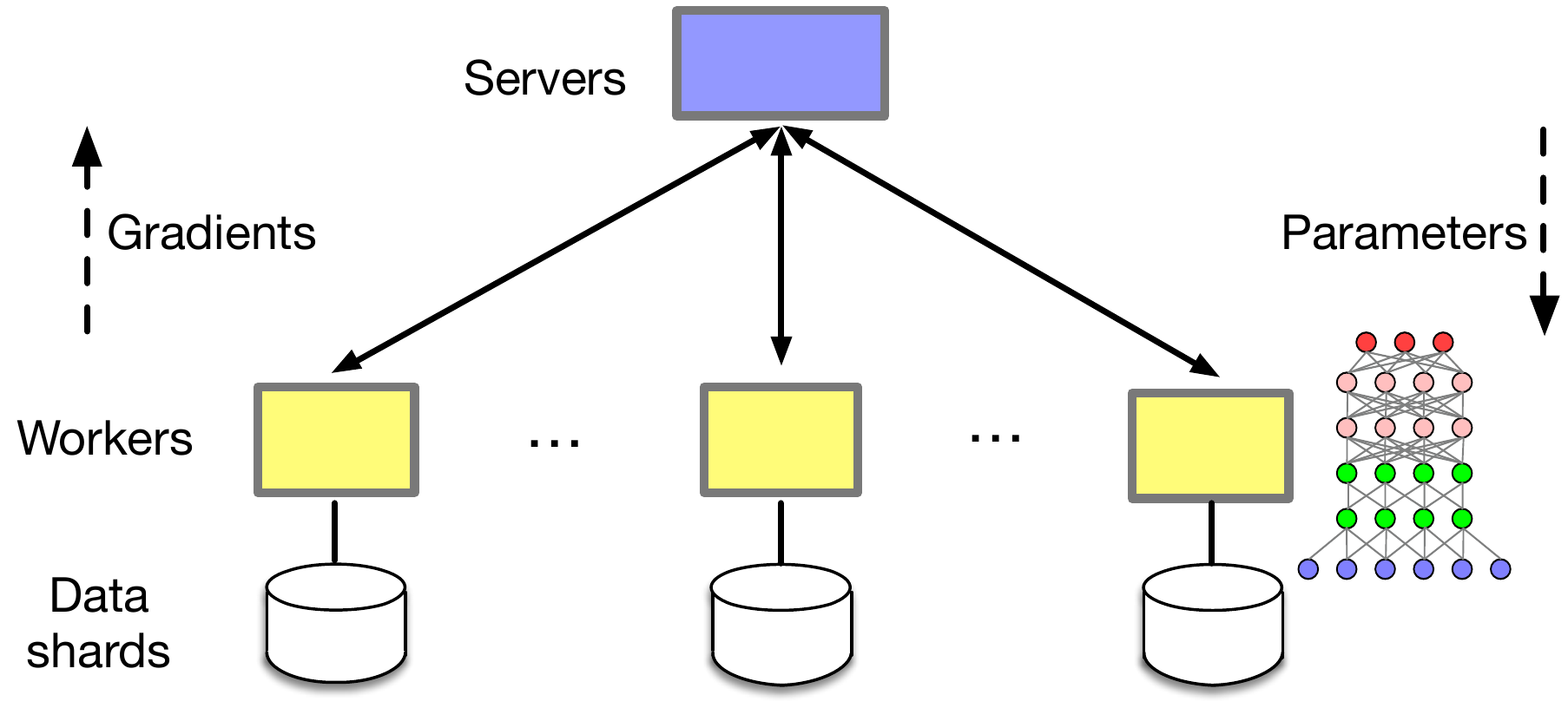}
\caption{The parameter server system. Each worker trains the complete
model using its local data.}
\label{fig:ps}
\vspace{-2mm}
\end{figure}

The parameter server (PS) system is a centralized design as shown in
Figure~\ref{fig:ps}. There are two types of nodes, {\em servers} and {\em
workers}. A server node maintains a partition of the globally shared model
parameters. Server nodes communicate with each other to replicate and/or to
migrate parameters for reliability and scaling.% A server manager node maintains a consistent view of the metadata of the servers, such as node liveness and the assignment of parameter partitions.
A worker node performs training using {\em data parallelism} \cite{ZWSL10}, and
is responsible for one partition or shard of the training dataset. Each worker
goes through a {\em mini-batch} (or simply batch) of its local data shard in
parallel to compute gradients as discussed in \cref{sec:dl}, and push them to
the corresponding servers. Servers then aggregate the gradients from all
workers to perform a global update of the model, and disseminate the new
parameters to each worker. This completes one {\em iteration} of training. When
workers go through all the samples in their shards in multiple iterations, they
complete one {\em epoch} of training. {Servers and workers are usually distinct
machines for better fault-tolerance and performance. }

\begin{comment}
The parameter server architecture provides implementation of basic engineering components of DL systems such as consistency models, fault-tolerance, and scalability, thereby enabling users to focus on application-specific code \cite{li2014scaling,chilimbi2014project} \hx{cite more}.
There is a scheduler node that assigns training tasks and data to workers and monitors their progress. 
If workers are added or removed, or upon worker failures, it reschedules unfinished tasks. There are two situations for the fault-tolerance when the worker is failure: (1) If the failure worker can recovery locally, so the worker restart training process locally; (2) If the failure worker cannot recovery locally or there is a new worker joining the training, the worker should pull the entir model prelica from the parameter server and get a partition of the training data from file system, then start the training process.
\hx{this needs a little more detail on say fault-tolerance}
\end{comment}

% \begin{figure*}[ht]
% \centering
% \subfloat[Parameter server system. Each worker trains the complete model using local data.]{\label{fig:ps}{\includegraphics[width=0.5\textwidth]{PS_architect}}}\hfill
% \subfloat[\sys where the training of CONV and FC layers are separated.]{\label{fig:ours}{\includegraphics[width=0.5\textwidth]{Cou}}}
% \caption{\footnotesize A motivation example using a simple CNN with two CONV layers and two FC layers.}
% \label{fig:motivation}
% \end{figure*}

\subsection{Sizable Data Exchange in Training}
\label{sec:bottleneck}

Most DL models contain a large number of parameters in order to capture the
complex features of the input data and their impact on the prediction results.
Further, each worker needs to push the gradient of every parameter, and pull
every updated parameter from the servers in each iteration. As a result,
distributed training in a PS system entails a substantial amount of data
exchange between workers and servers. 

Table~\ref{table:cpuvsgpu} shows the number of parameters for common DL models
in image classification, and the corresponding data exchange volume per
iteration in a PS system with 4 workers and 1 server. 
{All workers send gradients to and receive parameters from the server.}
With tens or
hundreds of millions of parameters, the PS server has to transfer about 0.8GB
to 4.4GB data over the network. The bulky data transfer incurs non-negligible
overhead to training time performance. The impact can be considerable
especially as GPUs are now prevalently used to train large DL models. Owing to
the high memory bandwidth and massive number of cores, GPU accelerates the
matrix computation in DL tremendously, leaving the system more susceptible to
other overheads such as the network data transfer.

\begin{table}[htp]
\centering
\footnotesize
\resizebox{\columnwidth}{!}{
\begin{tabular}{|c|c|c|c|c|}
\hline
\multirow{2}{*}{Model} & \multirow{2}{*}{\# Para.} & \multirow{2}{*}{Data Size 
(MB)} & Measured & Estimated \\
&  &  & Training Time & Comm. Time \\ \hline
AlexNet \cite{krizhevsky2012imagenet} & 61.1M & 488.8 $\times$ 4 & 1.99s &
1.56s\\
\hline
VGG-16 \cite{simonyan2014very}  & 138M & 1104 $\times$ 4 & 4.93s & 3.53s \\ \hline
VGG-19 \cite{simonyan2014very} & 143M & 1144 $\times$ 4 & 5.15s & 3.66s \\ \hline
Inception-V3 \cite{szegedy2016rethinking} & 27M & 216 $\times$ 4 & 0.83s & 0.69s \\
\hline
ResNet-152 \cite{he2016identity} & 60.2M & 481.6 $\times$ 4  & 1.86s & 1.54s \\
\hline
\end{tabular}
}
\caption{Number of model parameters
of popular DL models and its impact on training time per iteration with
ImageNet-12 \cite{ILSVRC15} and p3.2xlarge instances in AWS EC2. Each instance
has a Nvidia Tesla V100 GPU and 10Gb bandwidth. Training time results are
averaged over 100 iterations. Our PyTorch based PS implementation is used here
with 1 server and 3 workers.
Data size includes both gradients and parameters. Comm. time is estimated by
dividing total data size over 10G. Other hyperparameters are explained in
\cref{sec:setup}.}
% We measured the training time. The communication time is the difference between the training time and computing time which we measured. We measured the statistics based on the general topology since we cannot know the topology when we deploy the testbed in public cloud}}
\label{table:cpuvsgpu}
\vspace{-4mm}
\end{table}

% \xr{Plenty of data has been generating and exchanging at the same time. Data contains two parts, gradients and parameters.} 
% As a result, in PS systems, training performance is often bottlenecked on communication over network for parameter update \xr{heavy data exchange.} 

To demonstrate the problem, we implement a PS system on PyTorch
\cite{paszke2017pytorch} and use an EC2 GPU cluster with the same setup above
(4 workers and 1 server). We use p3.2xlarge instances each with a
state-of-the-art Nvidia Tesla V100 datacenter GPU \cite{V100} and 10Gbps
bandwidth, and train the DL models listed in Table~\ref{table:cpuvsgpu}. More
details of the implementation and testbed setup can be found in the evaluation
section \cref{sec:eval}.
{We measure the per iteration training time by inserting timestamps
before and after each iteration in the Python code. Note operations in PyTorch
are executed immediately when the Python statements are invoked due to its
dynamic graph design \cite{paszke2017pytorch}; we do not add any extra
barriers. The results averaged over 100 iterations are shown in
Table~\ref{table:cpuvsgpu}. We also show the estimated communication time to
complete the data transfer at the single server, which is calculated simply by
dividing the data size by the 10G bandwidth.} % overlap communication with computation

% \footnote{Execution is greedy w.r.t. dynamical graph design of . Operations are
% executed \hx{what's the point of this?} We insert many timestamps on some start
% points of processes, such as data loading, forwarding, synchronization and so
% on.}

\begin{comment}
We measure the total training time and computation time in each iteration, and summarize the average results over 100 iterations in Table~\ref{table:cpuvsgpu}.
\xr{Execution is greedy w.r.t. dynamical graph design of PyTorch \cite{paszke2017pytorch}. Operations are executed immediately when the Python statements are invoked. We put many time markers on some start points of processes, such like data loading, forwarding, synchronization and so on. We can get the time cost directly and accuracy.}  
The computation time includes time to load images to the GPU and to compute gradients on each worker. 
The rest of training time is then spent on communication \xr{which is a period from the first gradients arriving at the server to last parameter sending to the worker. }
\end{comment}

Observe from Table~\ref{table:cpuvsgpu} that the estimated communication time
takes $\sim$70\%--80\% of the measured training time for all models.
% \xr{There should no overlap. Even there is overlap, it just hides some time instead of reducing. 
{Note that actual impact of communication time may be less severe due to
various optimization techniques such as overlapping communication with
computation.} Nonetheless, the results reflect that data transfer has a
critical impact on distributed training with GPUs in PS systems. Prior work
\cite{203269,XKZH16,xiaowen} has reported similar observations and the problem
has attracted increasing attention recently in both DL
\cite{LHMW18,wen2017terngrad} and systems \cite{203269,iandola2016firecaffe}
communities.

The impact of data transfer certainly can be mitigated with higher network
bandwidth, which would require additional time and investment for the
infrastructure overhaul. On the other hand the problem may also aggravate with
the rapidly improving GPU or special-purpose hardware (ASIC, FPGA) that
slashes computation time. We therefore ask, is the massive data transfer
unavoidable in distributed training? Can we take a more fundamental approach
to minimize the data transfer and the training time in turn without affecting
the convergence or accuracy of the model?

\begin{comment}
Observe from Table~\ref{table:cpuvsgpu} that for all DL models, single worker needs to exchange hundreds of megabytes with the parameter server. For some models such as VGG-16 and VGG-19, the size of data even beyonds one thousand megabytes.
The results demonstrates empirically that large data exchange in the PS systems has the critical influence on the training time, owing to the massive update traffic generated by GPUs. The large data exchange indirectly proves that the cluster for distributed deep learning also need high capacity network.
\end{comment}

\begin{comment}
Observe from Table~\ref{table:cpuvsgpu} that for all DL models, communication attributes to $>$80\% of total training time per iteration. For some models such as AlexNet, VGG-16, and VGG-19, communication takes more than 90\% of training time. 
The results demonstrates empirically that communication is the performance bottleneck in PS systems, owing to the massive update traffic generated by GPUs. 
In fact some prior work \cite{203269,XKZH16,xiaowen} has reported similar observations and the problem has attracted increasing attention recently in both DL \cite{LHMW18,wen2017terngrad} and systems \cite{203269} communities.
\end{comment}
Note that some ML frameworks such as MXNet \cite{chen2015mxnet} designate
each node as both a worker and a server \cite{XKZH16}. Each node is
responsible for $1/N$ of the model parameters. This can mitigate the bandwidth
bottleneck at the servers. However, the amount of parameters each node needs
to send and receive is still massive ($4 (N-1)/N$ parameters), which does not
fundamentally overcome the issue. It is estimated that for such a system, the
network bandwidth has to be at least 26Gbps in order for it not to become the
bottleneck when training AlexNet on Titan X GPU \cite{203269}, which is slower
than the state-of-the-art GPUs now. Shi et al. \cite{xiaowen} also report that
MXNet has relatively high communication overhead when scaling to multiple
nodes even with 56Gb Infiniband.

\begin{comment}
network communication is the performance bottleneck in distributed DL systems with GPU. 
The communication bottleneck is likely to aggravate as the capability of GPU or special-purpose hardware (ASIC, FPGA) is improving more rapidly than the network bandwidth evolution.
For example Tesla V100 which is used in our experiments before is released in 2017 for production DL workloads. 
It offers 14 teraFLOPs single precision \cite{V100} which is 1.5x faster than its predecessor P100 released a year earlier. 
The server interconnects and routers on the other hand are still running at 40G or 100G at best, which has been available on the market for many years. 
Efficient solutions are thus required to tame the amount of parameter exchange in current PS systems, without sacrificing convergence or model accuracy.
\end{comment}

%% file: motivation.tex
%!TEX root = main.tex

\section{Separating the Layers}
\label{sec:motivation}
% \hx{start motivation}

To answer our quest, we examine the characteristics of DL models, and propose
to {\em separately} train the layers in order to substantially reduce the data
transfer in distributed training. We focus on convolutional neural networks
(CNNs) which are arguably the most widely used class of DL models.

\subsection{Layers are Remarkably Different}
\label{sec:layers}

\begin{figure}[ht]
\centering
\subfloat[AlexNet]{\label{fig:AlexNet}{\includegraphics
[width=0.4\textwidth]{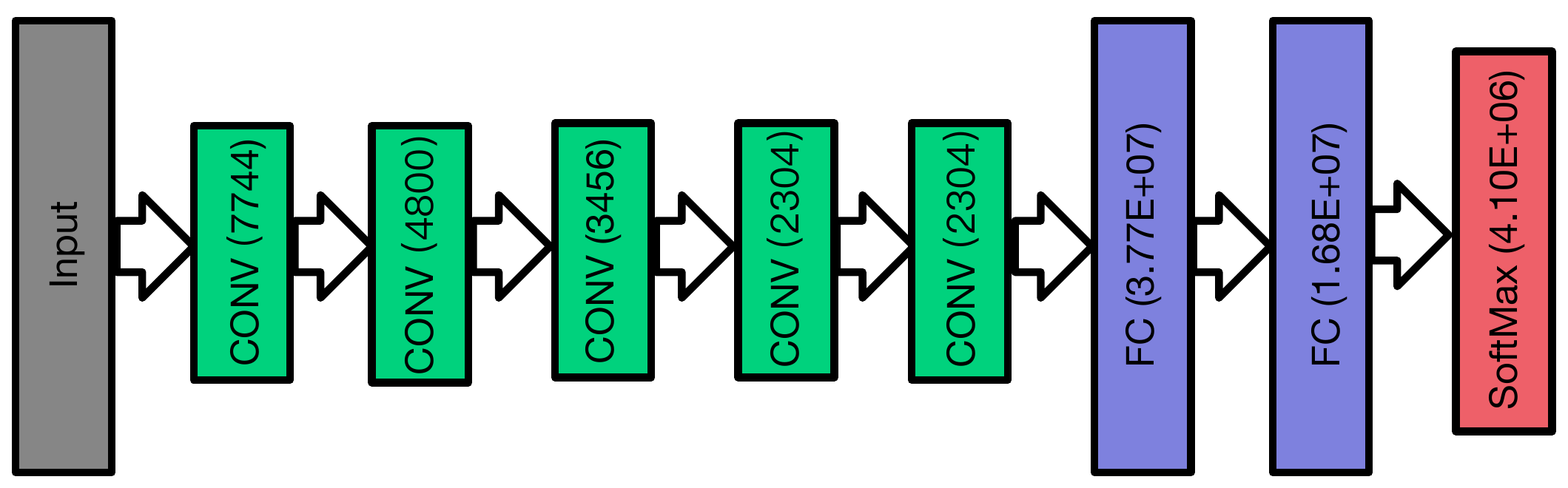}}}\hfill
\subfloat[VGG-16]{\label{fig:VGG-16}{\includegraphics[width=0.36\textwidth]
{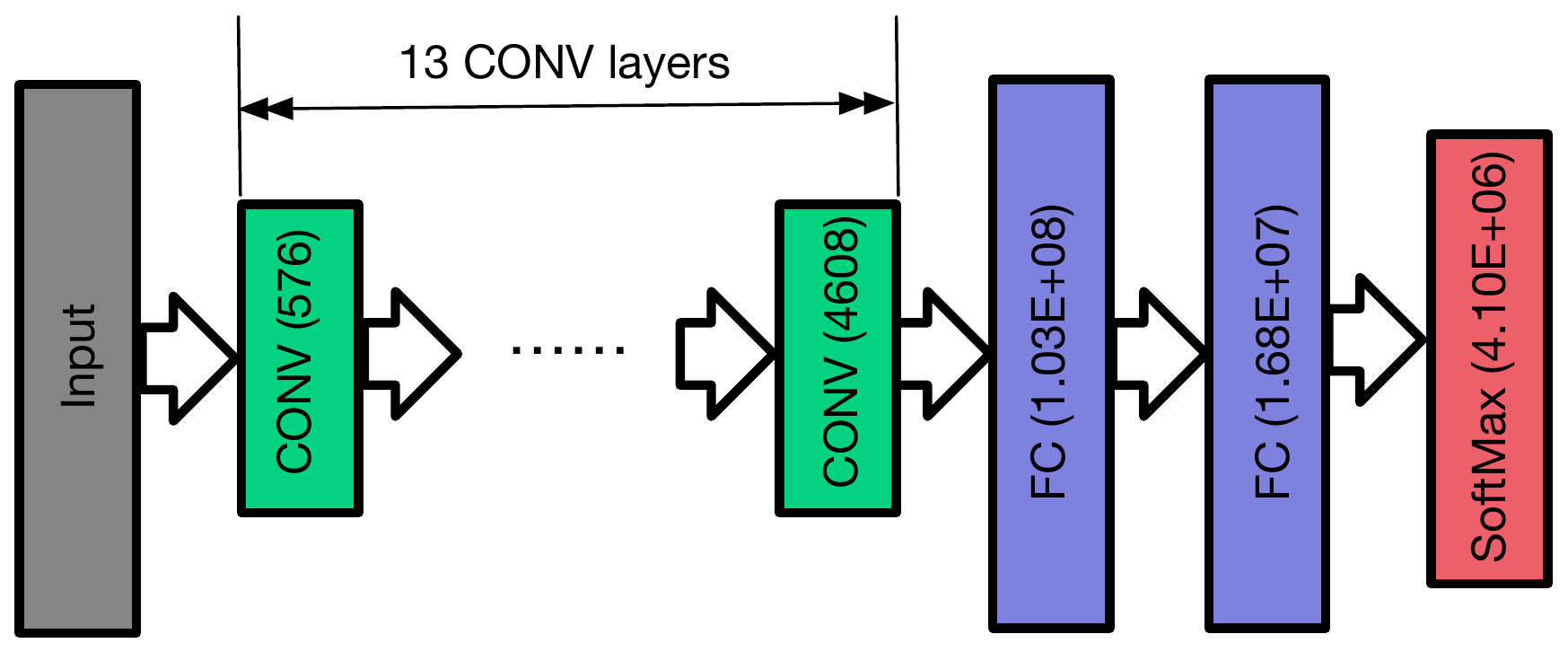}}}\hfill
% \vspace{-3mm}
\caption{Two common CNNs. AlexNet \cite{krizhevsky2012imagenet} has 5 CONV layers and 3 FC
layers; VGG-16 \cite{simonyan2014very} has 13 CONV layers and 3 FC layers. Other types
of layers such as the pooling layer do not contain parameters; they just
compute a fixed function of the inputs from the previous layer. Numbers of
parameters are indicated in parentheses.
}
\label{fig:cnn}
\vspace{-3mm}
\end{figure}

CNNs have been successfully applied in image recognition
\cite{krizhevsky2012imagenet,simonyan2014very,szegedy2016rethinking,he2016identity},
video analysis \cite{xing2015petuum}, natural language processing
\cite{kim2014convolutional}, drug discovery \cite{gawehn2016deep}, etc. A CNN
typically consists of two core building blocks, convolutional (CONV) layers
and fully connected (FC) layers
\cite{krizhevsky2012imagenet,simonyan2014very,szegedy2016rethinking,he2016identity}.
A CONV layer has a set of learnable filters each of which is spatially small
(e.g. 5$\times$5 along width and height). During the forward pass of training,
each filter is convolved across the input data to extract certain features at
some spatial position of the input. The neural network learns filters that
activate when they detect certain features (e.g. pointy ears, curled tails),
which are then passed on to the successive layers. Thus CONV layers are
usually positioned right after the input layer. After CONV layers, the
high-level reasoning is performed by the FC layers. Neurons in an FC layer
have connections to all outputs in the previous layer. FC layers employ
classifiers such as softmax to classify the output.

As a result of their functionality distinctions, CONV and FC layers exhibit
remarkably different characteristics. CONV layers have a small number of
parameters due to the small filters, but carry out a large amount of
convolution computations. On the contrary, FC layers usually have a very large
number of parameters due to its fully connected nature, but only require a
small amount of simple calculations. As shown in
Table~\ref{table:distribution} for example, for AlexNet and VGG-16, CONV
layers incur most of the computations ($>$90\%), while FC layers account for
most of the parameters ($\sim$90\%).\footnote{We count the number of
parameters which require gradients in the update. PyTorch provides API calls
to get this statistic. We cite the computation flops from \cite{ZHWX15}.}

\begin{table}[htp]
\centering
\footnotesize
\resizebox{0.7\columnwidth}{!}{
\begin{tabular}{|c|c|c|}
\hline
\# FLOPs & CONV Layers  & FC Layers  \\ \hline
AlexNet & 1.35E+09 / 92.0\% & 1.17E+08 / 8.0\% \\ \hline
VGG-16 & 1.09E+10 / 98.9\% & 1.21E+08 / 1.1\% \\ \hline
% AlexNet & 2.8E+09 / 97.9\% & 5.9E+7 / 2.1\% \\ \hline
% VGG-16 & 1.55E+10 / 97.3\% & 4.32E+08 / 2.7\% \\ \hline
\end{tabular}
}
\\
\vspace{3mm}
\resizebox{0.78\columnwidth}{!}{
\begin{tabular}{|c|c|c|}
\hline
\# Parameters & CONV Layers & FC Layers \\ \hline
AlexNet & 2.47E+06 / 4.04\% & 5.86E+07 / 95.96\% \\ \hline
VGG-16 & 1.47E+07 / 10.60\% & 1.24E+08 / 89.4\% \\ \hline
\end{tabular}
}
\vspace{3mm}
\caption{Estimated numbers of FLOPs and model parameters in AlexNet
\cite{krizhevsky2012imagenet} and VGG-16 \cite{simonyan2014very}. Data source:
\cite{ZHWX15}.}
\label{table:distribution}
% \vspace{-4mm}
\end{table}

\subsection{Separating CONV and FC Layers }
\label{subsec:process}
The distinction between CONV and FC layers presents salient opportunities for
us to optimize the communication cost of PS systems.

\begin{figure}[ht]
\centering
{\includegraphics[width=1\linewidth]{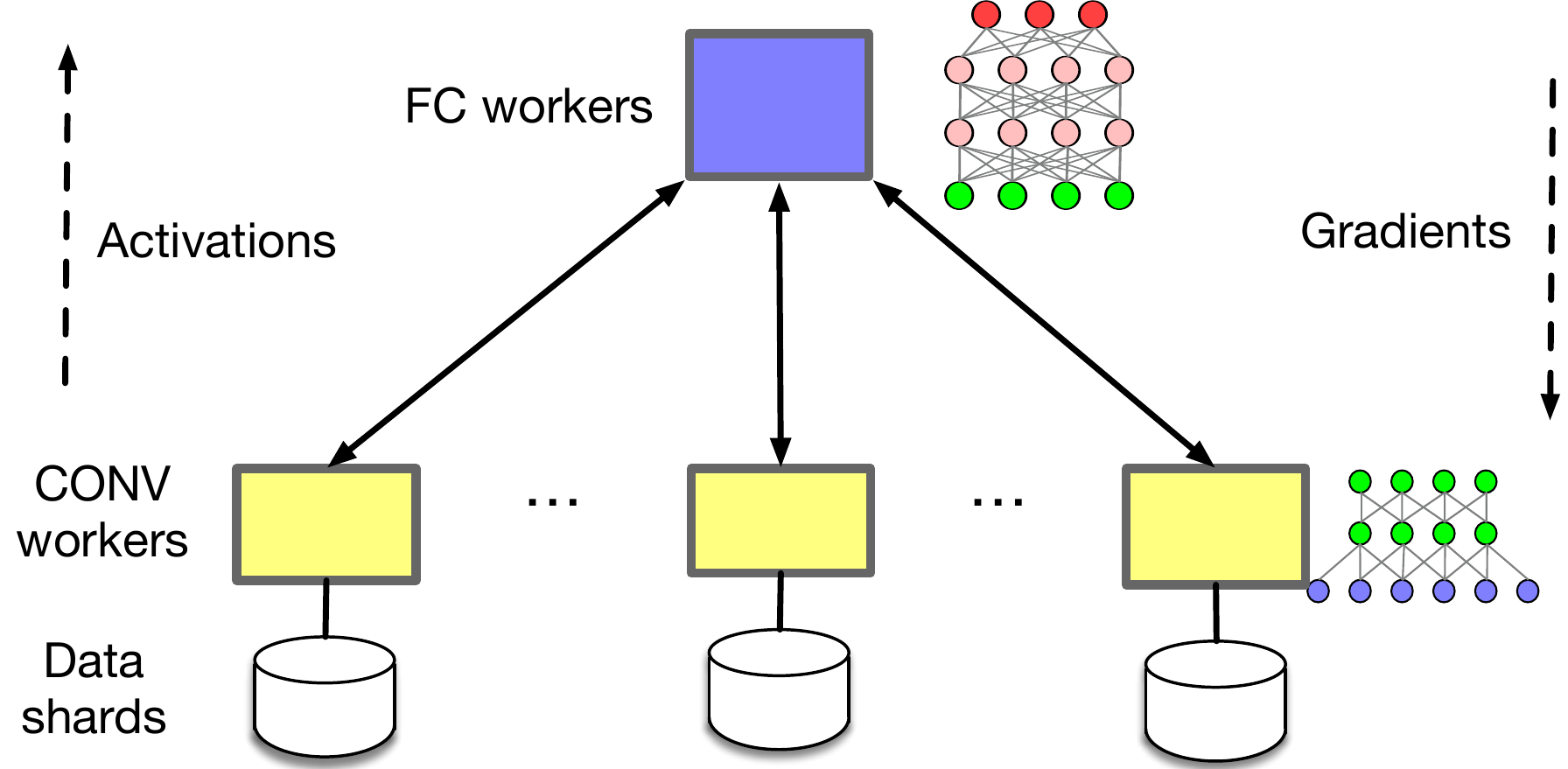}}
\caption{Training of CONV and FC layers are separated to reduce the
data transfer, particularly for the FC gradients and parameters.}
\label{fig:ours}
\end{figure}

Specifically, it is now clear that much of the data transfer between workers
and servers in a PS system is for FC layer parameters and gradients. % As discussed in \cref{sec:ps}, in PS systems each worker trains the entire model, and has to exchange gradients and parameters of the entire model with all servers. Thus much of the data transfer is in fact for FC layer parameters.
Since CONV layers require most computation, we can assign most of the machines
to train just the CONV layers, and the rest to train just the FC layers, as
shown in Figure~\ref{fig:ours}. With the separation of the layers, in the
forward pass the CONV workers send the output of the last CONV layer, i.e. the
{\em activations}, to the FC workers. The FC workers in turn send gradients of
the last CONV layer back to CONV workers to continue backpropagation. FC layer
gradients are now irrelevant to CONV workers, and no long need to be pushed
over the network to each worker. As a result, a significant part of the
communication in the traditional PS system can be eliminated to accelerate
training. This is our key idea.

\begin{figure}[ht]
\begin{centering}
\includegraphics[width=0.99\linewidth]{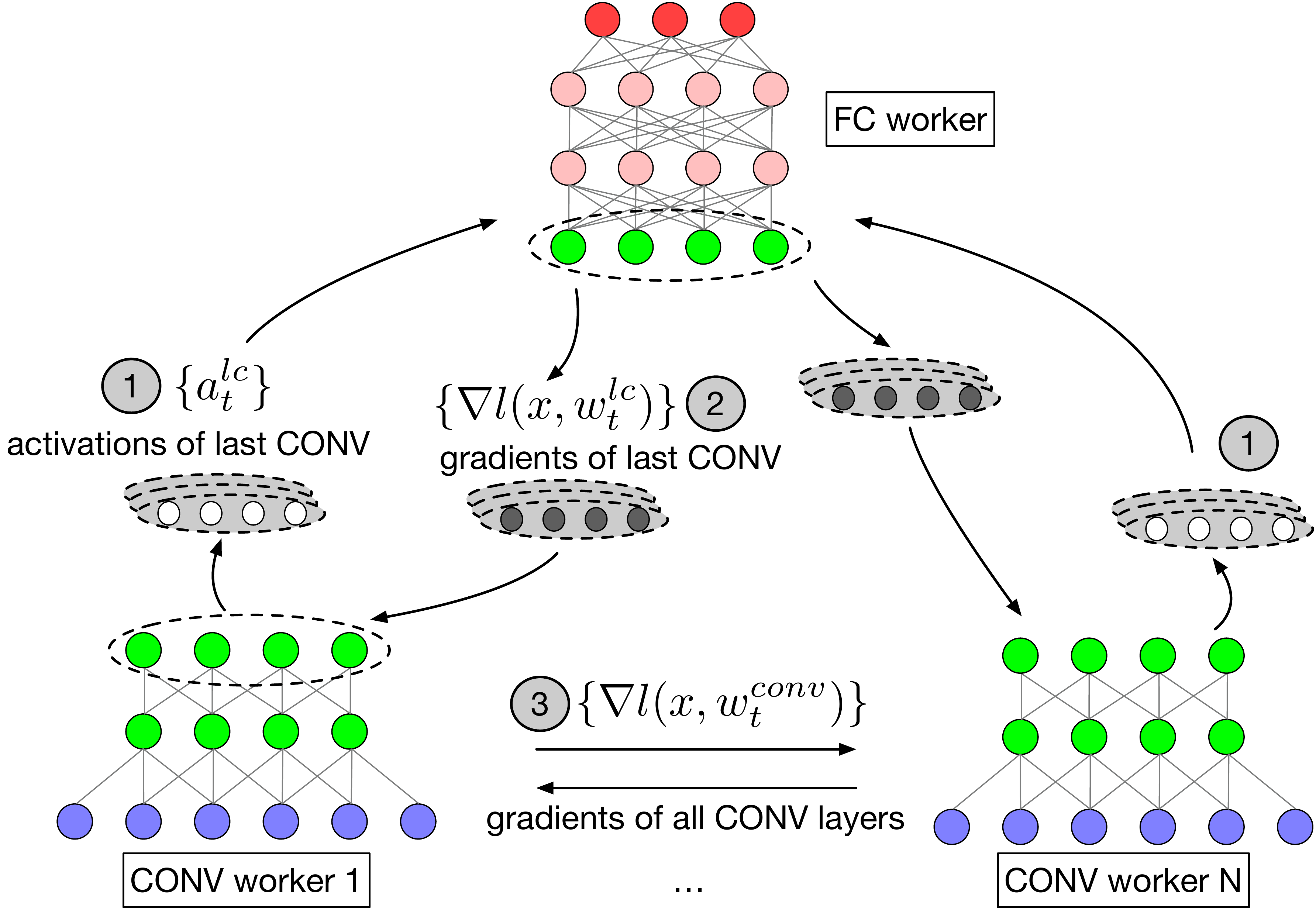}
\caption{The training process with the separation of CONV and FC layer training. At iteration $t$, CONV workers send activations of the last CONV layer $\{a^{lc}_t\}$ for a batch of samples to the FC worker. Then FC worker pushes gradients of the last CONV layer $\nabla l(x, w^{lc}_t)$. Lastly CONV workers exchange gradients of all CONV layers to update parameters.}
\label{fig:process}
\end{centering}
\end{figure}

{\bf Training Process with Layer Separation.}
Figure~\ref{fig:process} illustrates the training process with more detail.
Suppose there is one FC worker now since most computation concentrates on
CONV layers. In each iteration $t$, each CONV worker processes a sample up to
the last CONV layer, and sends the activations from the last CONV layer to the
FC worker. The FC worker collects all activations $a^{lc}_t$ for a batch of
samples, and continues the forward pass to compute the loss values. It then
starts backpropagation and computes gradients for all FC layers and the last
CONV layer. It pushes gradients of the last CONV layer $\nabla l(x, w^{lc}_t)$
to the CONV workers, and updates FC layer parameters. CONV workers then
continue the backpropagation to compute their gradients in parallel, exchange
them among all CONV workers to get global information $\sum_{x \in D_t} \nabla
l(x, w^{conv}_t)$ from all samples of the batch $D_t$, and perform CONV layer
parameter update accordingly. This completes one iteration of training. The
process is effectively equivalent to SGD in PS systems as explained in
\cref{sec:dl}.\footnote{Gradients from all CONV (resp. FC) workers are
disseminated to each CONV (resp. FC) worker.}

{Note that a (max) pooling layer may follow the last CONV layer to reduce the number of activations. As discussed pooling layer does not have any parameters. Thus in this case we separate the model by the last pooling layer and include it in CONV workers to reduce communication between FC and CONV workers. This does not change the training process.}

{\bf Benefits Overview.}
Now compared to PS, we eliminate the FC layer parameter exchange at the expense of: (1) sending activations and gradients of the last CONV layer, and (2) exchanging CONV layer gradients among CONV workers.
First, comparing with the number of parameters of the entire model, the number
of activations and gradients of the last CONV layer is small.
For example, the number of activations from the last pooling layer in AlexNet is just 9216 for each sample \cite{krizhevsky2012imagenet}. With a typical batch size of 128, a CONV worker sends a {total of $\sim$1.2M activations} compared to 61.1M parameters for the entire model in each iteration.
Second, the number of CONV layer parameters (gradients as well) is much smaller than that of {large} FC layers as discussed in \cref{sec:layers}, and we can optimize the exchange with parallel communication to further reduce the time (\cref{sec:hybrid_Comm}).
Thus, training time can be significantly improved by separating the CONV and FC layers. 

Essentially, our idea of separating the training of CONV and FC layers, and using activations to replace gradients, can be regarded as a mixed use of both data parallelism and model parallelism in the DL literature \cite{dean2012large}. 
Data parallelism---where each node trains the complete model with different data
partitions---is prevalent in current DL systems due to implementation
simplicity.
We also employ data parallelism among CONV workers which train the same CONV layers independently with different data partitions. 
At the same time, training of FC layers is done on a separate group of workers which is an example of {\em model parallelism}. 
Amid the discussion about the two paradigms in the community \cite{wu2015deep,dean2012large}, our hybrid approach represents a promising alternative to the common conception of using one in lieu of the other exclusively.

%% file: design.tex
%!TEX root = main.tex

\section{Design}
\label{sec:design}
\begin{comment}
1. how to divide the layers? just conv/FC
2. how many workers for FC? just 1
3. how do the conv workers comm.? 
\end{comment}

To demonstrate the feasibility of our idea, we design a new distributed DL
system called \sys. With the separate training of CONV and FC layers, \sys
needs to address two new challenges in system design: First, how to design the
communication strategies between CONV and FC workers, and amongst CONV workers
themselves in the backpropagation phase? Particularly, the CONV workers now
need to synchronize parameters among themselves, and it is critical that \sys
minimizes such overhead. Second, how to assign the nodes to perform CONV or FC
training, in order to ensure that the system performance is optimized in terms
of training throughput?
% \xr{After getting the approximately optimal deployment policy, \sys needs to show how to train via new architecture.} 

% We present the design of \sys in this section by discussing how it address these two new challenges.

\subsection{Hybrid Communication}
\label{sec:hybrid_Comm}

\sys utilizes hybrid communication strategies to address the first design challenge we just presented. 
% suit the requirements of different types of communications and minimize the impact of network transmission. 
For simplicity, we start the discussion assuming only one FC worker in the system. 
We discuss the case of multiple FC workers in \cref{sec:FC_comm}. 
With one FC worker, communication in \sys happens (1) between the CONV workers and the FC worker, and (2) among CONV workers themselves.

\subsubsection{CONV-FC Communication}
\label{sec:conv-fc_comm}
% \noindent{\bf CONV-FC communication.}

\begin{figure}[h]
\centering
\includegraphics[width=0.85\linewidth]{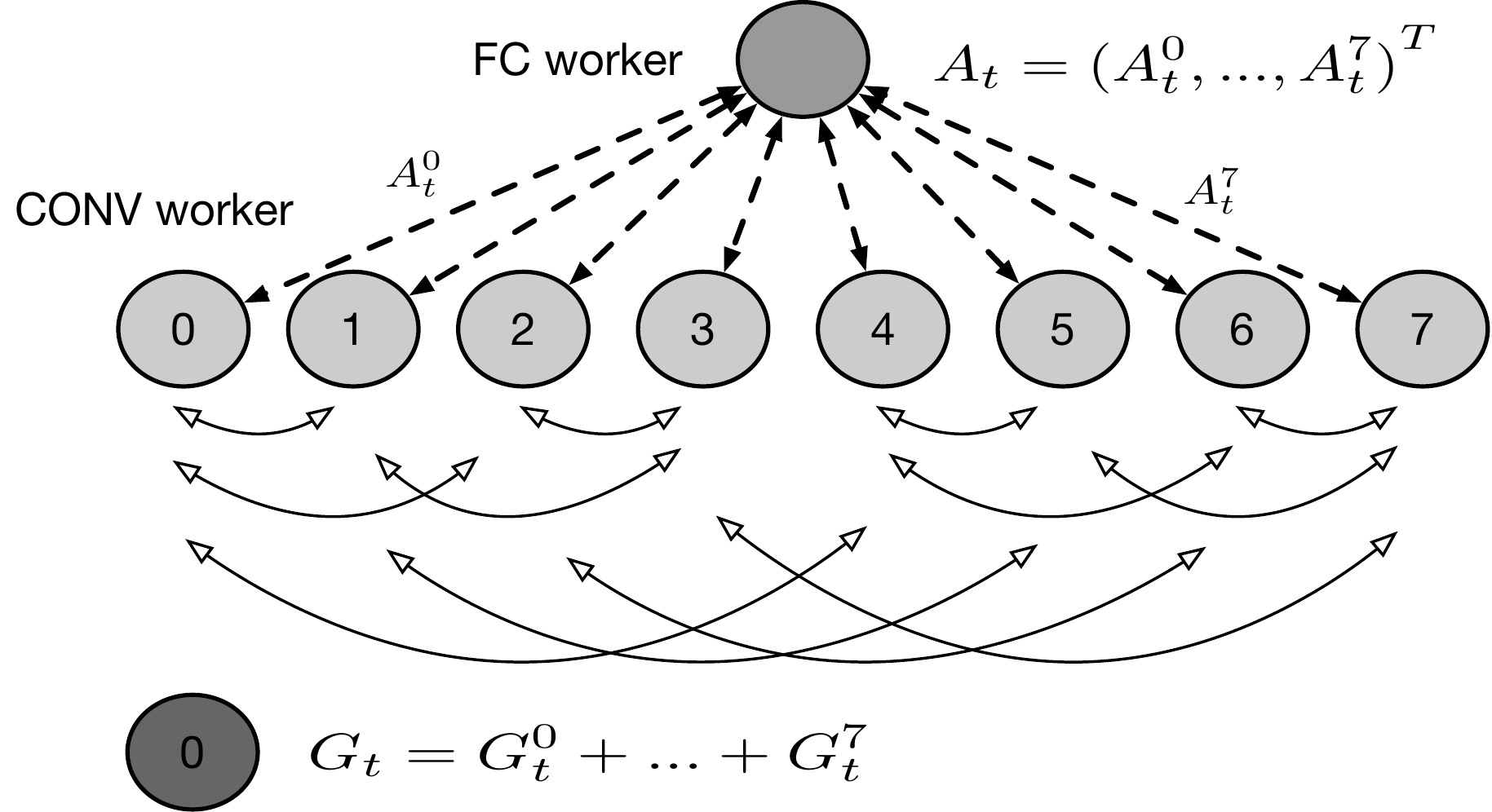}
\caption{Hybrid communication in \sys. Dashed lines represent the many-to-one
transmission from CONV workers to the FC worker, and one-to-many from FC worker to CONV workers. Solid lines represent the {\tt allreduce} communication among CONV workers in backpropagation, which is orchestrated by the recursive doubling algorithm \cite{thakur2005optimization}.}
\label{fig:HyComm}
\end{figure}

Activations of CONV workers are different since they process different input data. 
Thus gradients are also distinct for each CONV worker and has to be sent to the corresponding CONV worker in order for backpropagation to work. 
It is thus not possible to combine the activations or gradients to reduce the transmission volume. 
The number of activations and gradients of the last CONV layer is very small (a few MBs with a batch size of 128) and does not consume much bandwidth, as explained in \cref{subsec:process}. 
Therefore, \sys just uses simple many-to-one and one-to-many transmissions for CONV-FC communication as shown in Figure~\ref{fig:HyComm}. This is also easy to implement.
% A CONV worker sends activations to the FC worker right after finishing its forward pass, and a FC worker sends gradients to the corresponding CONV workers 

\subsubsection{CONV Worker Communication}
\label{sec:conv_comm}

CONV worker communication is more demanding since most nodes in the cluster are CONV workers. 
The number of gradients of all CONV layers is also larger than activations: AlexNet for example has 2.3M CONV layer parameters compared to 1.2M activations with a batch size of 128. 
Moreover, performing the gradient exchange in an all-to-all manner generates a large volume of data transfer that scales quadratically with the number of workers, and is clearly not efficient.
Note that the objective of CONV worker communication is to disseminate the
global average gradients among all CONV workers, which is essentially an {\tt allreduce} operation in MPI of parallel computing.
Therefore we adopt an efficient implementation of {\tt allreduce} called the
recursive doubling algorithm  \cite{thakur2005optimization} to schedule the communication.\footnote{Other more efficient implementations \cite{thakur2005optimization} cannot be adapted here because of non-commutative operations on tensors in PyTorch (operation cannot be commutative because this could introduce different numerical errors on different workers).} 
In recursive doubling, neighboring workers exchange data in the first round. Then workers that are two nodes away from each other exchange the {\em reduced} data from the first round. Generally workers exchange cumulative data with nodes $2i-1$ spots away in round $i$, as illustrated in Figure~\ref{fig:HyComm}.
In this case it requires $\log_2 8$, i.e. 3 rounds to complete {\tt allreduce}
with 8 CONV workers.
% When the number of processes is a power of two, it requires log2 P steps to reach the end of the algorithm

If the number of CONV workers $n_c$ is not a power of two, we first randomly pick $n_c - 2^{\lfloor\log_2 n_c\rfloor}$ ``surplus'' nodes and have them send their gradients to another $n_c - 2^{\lfloor\log_2 n_c\rfloor}$ nodes randomly in parallel.
Then recursive doubling can be applied on the $2^{\lfloor\log_2 n_c\rfloor}$ nodes without the surplus ones.
Finally the result is sent to surplus nodes in one round in parallel by $n_c - 2^{\lfloor\log_2 n_c\rfloor}$ random nodes. 
In other words it takes $\lfloor\log_2 n_c\rfloor + 2$ rounds to finish the {\tt
allreduce}.

\subsubsection{FC Worker Communication}
\label{sec:FC_comm}

Finally in the case of multiple FC workers, each is responsible for an equal
number of the CONV workers. Many-to-one and one-to-many communication
strategies are then used between the corresponding FC and CONV workers.
Each FC worker maintains a complete set of FC layers.
Among the FC workers, \sys relies on the same recursive doubling algorithm
described above to exchange gradients and parameters efficiently. Further, FC
worker communication is overlapped with CONV worker communication to reduce
training time, since FC workers can start the parameter exchange after sending
gradients to CONV workers.

\subsection{Node Assignment}
\label{sec:node_assignment}

% We explained the efficient communication strategy in the last section and only used one FC worker. In this section, 
We now proceed to discuss how \sys allocates the number of FC and CONV workers.  

There are many tradeoffs involved in node assignment. For example, the number of CONV workers determines the total batch size: more CONV workers enables more images to be trained in parallel at each iteration. At the same time, more activations have to be sent to the FC workers who also need to send back more gradients as well, and the communication among CONV workers also takes longer to complete. 
% \hx{can be strengthened}
Thus we rely on a principled approach to tackle this challenge:  
First we develop a performance model to characterize the key tradeoffs and their impact on \sys's training throughput, given a particular node assignment scheme. 
Then the optimal node assignment is obtained by solving the throughput maximization problem given the total number of nodes.

\subsubsection{A Performance Model}
\label{sec:perf_model}

We consider a homogeneous setting where each node has identical hardware resources.
The key notations in our model are summarized in Table~\ref{table:notations}. 
Lower case notations denote unknowns and upper case notations are known constants.
For tractability, our model only captures the ideal case performance without considering various overheads in for example node synchronization and parallelization, which are rather difficult to model explicitly. 
% We also include some notations for the parameter server architecture, which will be used in \cref{sec:simulation}. 

\begin{table}[h]
\centering
\footnotesize
\begin{tabular}{|c|l|}
\hline
$N$                & total number of nodes \\ \hline
% $n_w, n_s$         & number of workers and parameter servers        \\ \hline
$n_c, n_f$         & number of CONV workers and FC workers   \\ \hline
$K$                & per-node batch size                             \\ \hline
$P$                & total number of parameters in the DL model   \\ \hline
$A$                & number of activations of the last CONV layer \\ \hline
$P_c$              & number of parameters in all CONV layers \\ \hline
$B$                & bandwidth at each node    \\ \hline
% $\alpha$           & coefficient of $T_{f}$ depends on GPU and model     \\ \hline
$t^c_{s}$	     & computation time per iteration \sys   \\ \hline
$T_{c}$ 		   & \begin{tabular}[c]{@{}l@{}}single worker CONV layer computation time \\ per iteration  \end{tabular} \\ \hline
$T_{f}$				& \begin{tabular}[c]{@{}l@{}}single worker FC layer computation time \\ per iteration for one CONV worker's activations  \end{tabular} \\ \hline
$t_{s}$    & total training time per iteration for PS and \sys  \\ \hline
\end{tabular}
\caption{Notation in the performance model.}
\label{table:notations}
% \vspace{-4mm}
\end{table}

We first characterize the computation time per iteration in \sys $t^c_s$.
There are two components: (1) CONV computation time $T_{c}$, including time to compute activations in the forward pass, and time to compute CONV layer gradients in backpropagation. This term only depends on the per-node batch size and a node's GPU resources, and is constant with respect to node assignment since CONV workers work in parallel.
(2) FC computation time, including the time to finish the forward pass based on activations, and then generate gradients for the FC layers. 
Denote the FC computation time to handle activations from one CONV worker on one FC worker as $T_{f}$. 
It is also constant since $T_{f}$ only depends on the per-node batch size and a node's GPU resources.
With $n_c$ CONV workers, the FC computation time grows to $n_c T_{f}$ due to the increase in activations, and with $n_f$ FC workers it becomes $\frac {n_c}{n_f} T_{f}$ due to parallel processing.
Taken everything together, we have:
\begin{equation}
t^c_s = T_{c} + \frac {n_c}{n_f} T_{f}.
\end{equation}

Now consider communication time in each iteration, which includes (1) CONV-FC communication time, (2) FC worker communication time, and (3) CONV worker communication time. 
Notice that (2) and (3) overlap in time as explained in \cref{sec:FC_comm}.
% \xr{FC worker communication occurs between the time when the FC workers send back gradients to CONV workers and receive the activation from CONV workers in next iteration, the communication cost is almost hided in other cost.}
Thus we do not consider FC worker communication here.

CONV workers send $n_cAK$ activations of the last CONV layer to the FC workers, which in turns send back $n_cAK$ gradients. Thus, CONV-FC communication takes ${2n_cAK}/{n_fB}$.
CONV worker communications takes ${log_2 n_c}$ rounds if $n_c$ is a power of 2, and $\lfloor log_2 n_c \rfloor +2$ rounds if otherwise as explained in \cref{sec:conv_comm}.
Each round takes $P_c/B$. 
Therefore, training time per iteration in \sys, including communication and computation, can be written as:
\begin{equation} 
\label{eqn:MT}
t_s =
	\begin{cases}
	t_s^c + \frac{2n_cAK}{n_f B} + \frac{P_c log_2 n_c}{B}, &  \text{if } n_c \text{ is a power of 2,}\\
	t_s^c + \frac{2n_cAK}{n_f B} + \frac{P_c \left(\lfloor log_2 n_c \rfloor +2\right)}{B}, &  \text{otherwise.} 
	\end{cases}
\end{equation}

The performance of a distributed DL system is characterized by training throughput, i.e. how many samples can be processed in unit time. 
We know that the total batch size is $n_c K$ in \sys, i.e. with $n_c$ CONV workers it processes $n_c K$ samples in one iteration. 
Hence, its training throughput can be expressed as:
\begin{equation}
throughput_s(n_c, n_f) = \frac {n_c K}{t_s}.
\end{equation}

\subsubsection{Node Assignment}
\label{sec:assignment}

Observe from the performance model that a larger $n_c$ increases the total batch size, but also increases the computation time $t_s$ and communication time. 
Node assignment then strives to maximize the throughput of the cluster given the total number of nodes $N$.
Mathematically, the optimal node assignment solves the following program:
\begin{equation}
\begin{aligned}
& \underset{n_c,n_f} {\text{maximize}}
& & throughput_s(n_c,n_f) \\
& \text{subject to}
& & N = n_c + n_f, \\
&&& n_c > 0, n_f > 0. 
% &&& GRAMneed(n_c,n_f) \leq GPUram.
\end{aligned}
\label{opt:throughput}
\end{equation}
The optimization program can be solved offline by an exhaustive search over all
possible values of $n_c$, as the total number of nodes is at most hundreds or
thousands in practice and the computational cost is very small. More discussion
on searching for the optimal deployment policy is in \cref{sec:dis}.
Note that for a given CNN model, the constants $T_c$ and $T_f$ depend on the node's GPU resources and can be profiled offline fairly accurately.

%% file: implement.tex
\section{Implementation}
\label{sec:implementation}

We implement \sys based on the popular DL framework PyTorch \cite{paszke2017pytorch} and its distributed communication package with TCP as the backend.
Our prototype has two main components: a controller that maintains the model and cluster configuration, and a communication library that can be used in PyTorch programs to handle parameter communication for both PS and \sys. 
The software architecture is similar to prior implementation of distributed DL training \cite{203269}. 

\begin{table*}[ht]
\centering
\footnotesize
\begin{tabular}{|l|l|l|}
\hline
Method           & Owner                & Description \\ \hline
% Decompose            & model                       & Divide the model into CONV part and FC part                                                                                                                                \\ \hline
{\tt push} 		& Worker in PS	 & Send gradients to the corresponding servers in PS                                                                                                                                      \\ \hline
{\tt pull} 		& Worker in PS	 & Receive updated parameters from the corresponding servers in PS                                                                                                                                      \\ \hline
{\tt push\_activation} 		& CONV worker	& Send activations of the last CONV layer to FC workers in \sys \\ \hline

{\tt pull\_fc} 		& CONV worker	 & Receive gradients of the last CONV layer from the corresponding FC workers in \sys \\ \hline
{\tt pull\_grad} 		& CONV/FC worker	 & Exchange gradients of all CONV/FC layers among CONV/FC workers in \sys \\ \hline
% GRAMneed()           & A list of training settings & Compute the GRAM needed to train the model                                                                                                                                 \\ \hline
\end{tabular}
\caption{Communication APIs in our PyTorch prototype for both PS and \sys. }
\label{table:api}
\vspace{-4mm}
\end{table*} 

\subsection{Controller}
\label{sec:controller}

The client program in PyTorch first instantiates our prototype by creating a controller within its process, and passing information about the entire CNN model and hyperparameters (e.g. batch size) to it. 
The program also specifies which architecture is to be used for distributed training, PS or \sys. 

For PS, the controller obtains the list of server and worker nodes, partitions the parameters equally across the servers by hashing, sends the mapping to each worker, and partitions the training data equally into data shards. 
It also sends the entire CNN model to workers to prepare for training.

For \sys, the controller collects information about the throughput maximization program \eqref{opt:throughput} in \cref{sec:assignment} such as available bandwidth, single worker CONV/FC computation times ($T_c$ and $T_f$), etc., and solves the problem to compute the number of FC and CONV workers needed. 
It sends all FC layers and the last CONV layer to the FC workers, and all CONV layers to CONV workers. 
In case of multiple FC workers, the CONV workers are equally partitioned into $n_f$ groups and the mapping is maintained and synchronized by each CONV worker.

\subsection{Communication Library}
\label{sec:comm_lib}

The communication library can be plugged into the training program. 
It provides APIs to support parameter communication with \sys and PS as shown in Table~\ref{table:api}. 
The {\tt push} and {\tt pull} methods are used by workers in PS to send gradients to servers, and acquire parameters from servers, respectively, based on the parameter mapping information. The {\tt pull} method is called immediately after {\tt push}, and blocks until it receives all parameter updates. The {\tt push} method is nonblocking. 
The {\tt push} and {\tt pull} are implemented using PyTorch's {\tt send} and {\tt receive} primitives, respectively, with the widely used bulk synchronous parallel (BSP) model \cite{chen2016revisiting}.

For \sys, the CONV workers invoke the {\tt push\_activation} method, which collects only activations from the last CONV layer for a batch of samples, then uses PyTorch's {\tt send} primitive to perform many-to-one communication to the corresponding FC workers.  
They then immediately call the blocking {\tt pull\_fc} method to obtain gradients of the last CONV layer from the corresponding FC workers.
Finally, the {\tt pull\_grad} method is called by each CONV worker to perform gradient exchange with the recursive doubling algorithm discussed in \cref{sec:hybrid_Comm}. It uses PyTorch's {\tt all\_reduce} primitive. It blocks until the operation is finished for all CONV workers.
It is also called by each FC worker to exchange gradients when there are more than one FC worker.
The APIs for \sys also apply BSP for model consistency.

% \begin{comment}
\subsection{Fault-Tolerance}
\label{sec:misc_imp}

For fault-tolerance, we use checkpointing throughout the system. 
Each node regularly creates checkpoints for current parameters and training state. 
In \sys, there are very few FC workers. Thus a FC worker will create additional checkpoints for its FC parameters in a randomly chosen CONV worker. The redundant checkpoint is sent when both CONV and FC workers are training the model on GPUs, creating little overhead on training time.

%% file: evaluation.tex
\section{Evaluation}
\label{sec:eval}

We conduct testbed experiments on GPU clusters to assess the performance of the \sys. 
We first verify its effectiveness with small datasets, by showing that \sys achieves the same training accuracy as PS using 2x to 3.8x less time. 
Then we demonstrate the performance benefit of \sys in production settings with large datasets and models, by showing that it achieves 1.34x to 13.9x speedup over PS, and the gain is more salient as the cluster grows.

\subsection{Setup}
\label{sec:setup}
\noindent\Emph{Testbed in Public Clouds.} 
To understand \sys's performance in a realistic environment, we deploy our PyTorch-based prototype on GPU clusters from two cloud providers, Azure and EC2. 
(1) We use {standard\_NC6} VMs from Azure as our first cluster. 
Each node is equipped with 6-core vCPUs, 56GB RAM, and a Nvidia Tesla K80 GPU (half of a physical card). Nodes are interconnected with high bandwidth, which we find to be $\sim$2.6Gbps.
(2) For the EC2 cluster we use p3.2xlarge instances.
Each node has 8-core vCPUs, 61GB RAM, and a Nvidia Tesla V100 GPU with 16GB RAM. The network bandwidth across nodes is 10Gbps. 
All nodes run Ubuntu~16.04, Nvidia driver version 384.13, CUDA 9.0, cuDNN 7.0, and PyTorch 0.3.1. 

\sys and PS always use the same number of nodes as CONV workers and workers, respectively, for fairness. 
\sys uses 1 FC worker in all settings here. This does not imply that node
assignment in \cref{sec:node_assignment} is not useful. Instead, one FC worker
is indeed optimal for throughput maximization when the cluster is smaller than
10 nodes in our testbeds.\footnote{We find that GPU memory of a single FC worker
is enough for
10 CONV workers.} For larger clusters we need to use multiple FC workers as
will be shown in \cref{sec:simulation}. PS also uses 1 server for fairness
unless otherwise stated. Thus the total batch size is identical for the two
systems, and we focus on training time per iteration as the  performance
metric here.

\noindent\Emph{Datasets.} We use two image classification datasets that are widely used in prior work \cite{li2014scaling,203269,coates2011analysis,abadi2016tensorflow,lian2017can}. 
(1) CIFAR-10 \cite{krizhevsky2014cifar}: It is a small dataset consisting of 60K
32$\times$32 color images in 10 classes, with 50K training images and 10K
validation images. 
(2) ImageNet-12 \cite{ILSVRC15}: This is a large dataset used in the annual
ImageNet Large Scale Visual Recognition Challenge (ILSVRC), with 1.28 million
training images and 50K validation images in 1K classes. 
% The datasets are stored in each node in advance. 

\begin{table}[htp]
\centering
\footnotesize
\resizebox{\columnwidth}{!}{
\begin{tabular}{|c|c|c|c|}
\hline
Model        & \# Params & Dataset     & Batch size \\ \hline
VGG-16 \cite{simonyan2014very}      & 33.6M     & CIFAR-10 \cite{krizhevsky2014cifar}   & 128       \\ \hline
AlexNet \cite{krizhevsky2012imagenet}     & 61.1M     & ImageNet-12\cite{ILSVRC15} & 128       \\ \hline
VGG-16  \cite{simonyan2014very}     & 138M      & ImageNet-12 & 64        \\ \hline
VGG-19 \cite{simonyan2014very}      & 143M      & ImageNet-12 & 64        \\ \hline
Inception-V3 \cite{szegedy2016rethinking} & 27M       & ImageNet-12 & 32        \\ \hline
ResNet-152 \cite{he2016identity}  & 60.2M     & ImageNet-12 & 32        \\ \hline
\end{tabular}}
\caption{Combinations of models and datasets used in the evaluation. Batch size is for each worker. 
}
\vspace{-3mm}
\label{table:models}
\end{table}

\noindent\Emph{Models and Hyperparameters.}
We evaluate \sys across many common CNNs for image classification. 
\begin{comment}
(1) AlexNet \cite{krizhevsky2012imagenet}: This is the model that won the ImageNet Challenge 2012 and subsequently spurred much attention on deep learning. AlexNet has only 8 layers, the first 5 are CONV layers, and the last 3 FC layers. 
(2) VGG-16 and VGG-19 \cite{simonyan2014very}: Proposed in 2014, VGG models are deeper than AlexNet to extract more features from images. 
VGG-16 has 13 CONV layers and 3 FC layers, and VGG-19 has 3 more CONV layers. 
(3) Inception-V3 \cite{szegedy2016rethinking}: This is also a winner of the ImageNet Challenge in 2016. 
It uses the Inception module, a mini-model that replaces large convolutions to reduce the computations and parameters. 
(4) ResNet-152 \cite{he2016identity}: This is a very deep CNN with 152 layers. This model adopts residual blocks which can ease the training of deep networks.
\end{comment}
We list the model configurations in Table~\ref{table:models}.
We use the momentum SGD \cite{goyal2017accurate} as the optimization algorithm in both \sys and PS. 
Small weight decay and dropouts are used to get better performance. 

% The total batch size across all nodes needs to be below 1024, for otherwise training may not converge as reported in \cite{goyal2017accurate}. Thus we set the per-node batch size to 128 for AlexNet on ImageNet-12 and VGG-16 on CIFAR-10 for we have at most 8 workers. In all other cases we adopt the recommended batch size configurations from the original papers \cite{krizhevsky2012imagenet,simonyan2014very,szegedy2016rethinking,he2016identity}.

\subsection{Baseline Performance }
\label{sec:baseline}

We first evaluate the baseline performance of \sys using the small CIFAR-10 dataset and VGG-16. 
% CIFAR-10 is a small dataset commonly used for quick verification. 
We modify the last softmax layer in the original VGG-16 in order to use CIFAR-10 with less classes, which reduces the number of parameters to 33.6M as shown in Table~\ref{table:models}. 
Since the dataset and model are small, the experiments are done on the Azure cluster with less bandwidth. 
% One node is used as the server in PS or as the FC worker in \sys, and the number of (CONV) workers varies in different settings. 
The same hyperparameters in Table~\ref{table:vgg-cifar10} are used.
Consistent with \cite{simonyan2014very} we decrease the learning rate by 0.1 every 30 epochs. 
We train the model for 40 epochs.
% \hx{which is long enough to achieve convergence}.  
Recall that as in \cref{sec:ps}, one epoch is one pass of the full training set and consists of many iterations during which each worker processes one batch of samples. 

\begin{table}[h]
\centering
\footnotesize
\resizebox{\columnwidth}{!}{
\begin{tabular}{cccccc}
\hline
\# (CONV) & Base & Total & \multirow{2}{*}{System}              & Training  & \multicolumn{1}{l}{Accuracy} \\
Workers 	& LR 	& batchsize 	& 	& time (h)\\ \hline
\multirow{2}{*}{2} & \multirow{2}{*}{0.2} & \multirow{2}{*}{256}  & \sys & 1.78              & 92.21\%                         \\
                   &                      &                       & PS                  & 3.64              & 90.95\%                             \\ \hline
\multirow{2}{*}{4} & \multirow{2}{*}{0.4} & \multirow{2}{*}{512}  & \sys & 1.22              & 91.52\%                            \\
                   &                      &                       & PS                  & 3.76              & 90.54\%                            \\ \hline
\multirow{2}{*}{8} & \multirow{2}{*}{0.8} & \multirow{2}{*}{1024} & \sys & 0.93              & 89.80\%                            \\
                   &                      &                       & PS                  & 3.52              & 89.95\%                             \\ \hline
\end{tabular}}
\caption{Baseline performance of \sys with VGG-16 on CIFAR-10 using the Azure
GPU cluster. We use 2, 4, and 8 CONV workers for \sys and the same numbers of
workers for PS, respectively. Per-node batchsize is 128. We use the same
hyper-parameters and deployment policies for \sys and PS in the same setting. We
train for 40 epochs in each setting.} 
\label{table:vgg-cifar10}
\vspace{-4mm}
\end{table}

Table~\ref{table:vgg-cifar10} shows the performance.  
\sys effectively achieves the same (slightly better actually) accuracy of $\sim$90\% compared to PS in all cases after 40 epochs. 
Further, \sys brings significant savings in training time (at least 2x speedup), and the savings are more salient when the cluster grows. 
With 8 workers, \sys obtains about 3.8x speedup compared to PS. 

We also note that as the cluster grows the total training time for PS barely improves. 
The main reason is that network bandwidth is the bottleneck here, and though adding workers reduces the number of iterations per epoch, the amount of gradient and parameter transfer to/from the server also increases proportionally in PS. The total amount of gradient and parameter exchange per iteration stays the same as a result.
We measure the time spent on computation in PS over 40 epochs, which is 0.47h and 0.25h for 4 and 8 workers, respectively. The numbers are reasonable, and they imply that the communication time is 3.29h and 3.27h with 4 and 8 workers, respectively.\footnote{
The results are in line with our settings in Table~\ref{table:vgg-cifar10}. For instance with 8 workers in PS, one epoch has 50000/8/128=48 iterations (the remainder samples are dropped). With 2.6Gbps bandwidth we can calculate the communication time to be 33.3M*4*2*8*48*40*8/1024/1024/1024/2.6/3600=3.25h.} 
% This implies that the computation time is , which is reasonable.}

\noindent{\bf Quick recap:}
The experiments here verify \sys's effectiveness: It improves training time by up to 3.8x compared to PS without trading off training accuracy, in settings with limited bandwidth. 

\begin{figure*}[t]
\centering
\subfloat[Speedup]{\label{fig:1ps_speedup}{\includegraphics[width=0.33\textwidth]{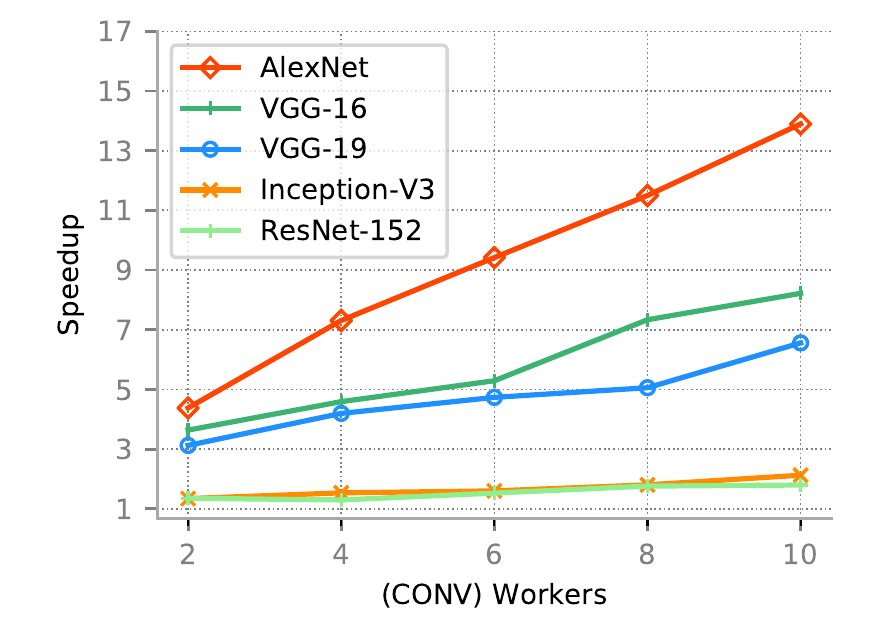}}}\hfill
\subfloat[FC-Data]{\label{fig:1ps_fc}{\includegraphics[width=0.33\textwidth]{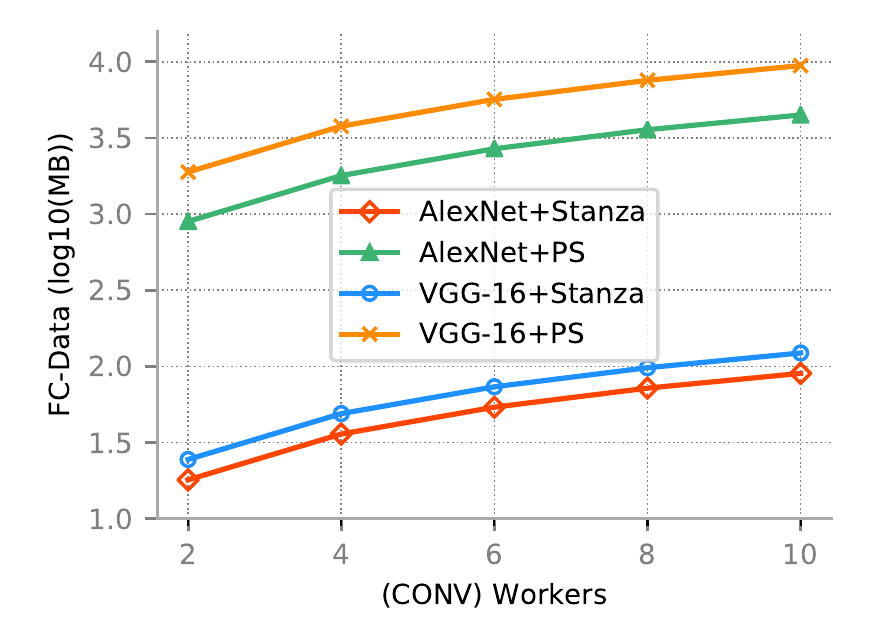}}}\hfill
\subfloat[Total-Data]{\label{fig:1ps_max}{\includegraphics[width=0.33\textwidth]{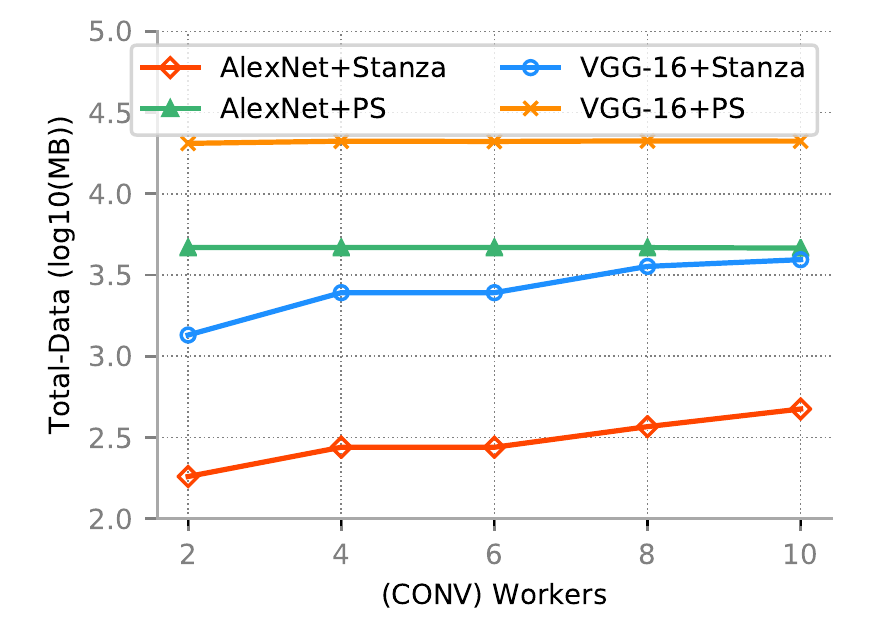}}}
% \vspace{-3mm}
\caption{Performance comparison with ImageNet-12 on large CNNs using the EC2 GPU cluster. PS uses one server and \sys uses one FC worker for fairness. FC-Data and Total-Data are shown in log scale.}
\label{fig:1PS_COM}
% \vspace{-3mm}
\end{figure*}

\subsection{Performance for Large CNNs }
\label{sec:perf_large}

We now use large CNNs to evaluate \sys with the ImageNet-12 dataset. 
The experiments are done on the EC2 cluster with Tesla V100 GPUs and 10Gb bandwidth, which represent typical production settings for training DL models.
One node is used as the server in PS and FC worker in \sys, and up to 10 nodes are used as workers in PS and CONV workers in \sys.
% \xr{We fix the stable single-worker batchsize in Table~\ref{table:models}, and linear scaling the learning rate with the size of cluster.}
The same hyperparameters in the previous section are used here, and the learning rate is linearly scaled with the cluster size. 
{Training lasts for 30 epochs and both PS and \sys achieve same accuracy, and we omit the accuracy results for brevity.}

We focus on three performance metrics now: 
(1) Speedup: This is the ratio between training times of PS and \sys. 
Larger speedup indicates better training time improvements. 
(2) FC-Layer Data Transfer (FC-Data): This is the amount of data transfer needed (in MB) to update the FC layer parameters in each {\em iteration}. 
Since we use 1 FC worker, FC-Data for Stanza effectively includes the number of activations and corresponding gradients only. 
(3) Total Data Transfer (Total-Data): Total-Data is defined as the total amount of data transfer (in MB) across all nodes in the system in an {\em epoch}. 
We use different time granularities for Total-Data and FC-Data to provide more insights into the performance analysis.
Note that with the ImageNet-12 dataset an epoch has many iterations now. For example from Table~\ref{table:models}, an epoch has 1251 iterations when we train AlexNet with 8 workers and a per-worker batch size of 128. 

Figure~\ref{fig:1PS_COM} depicts the performance results with a varying number of (CONV) workers. 
We make two interesting observations here.
First \sys consistently shows salient speedups from 1.34x to 13.9x over PS as shown in Figure~\ref{fig:1ps_speedup}. 
Among all models, AlexNet has the best speedup, and Inception-V3 and ResNet-152 the smallest. 
This is because Inception-V3 and ResNet-152 have proportionally more parameters in the CONV layers than other models, so the benefit of eliminating FC layer parameter exchange is not as substantial. 
% \xr{Inception-V3 and ResNet-152 have less parameters in the FC layers, though activations can compress 10x-30x data to update parameters in FC layers, the benefit for the entire communication is small.} \hx{well ResNet-152 has same number of parameters as AlexNet. so explain more here}

Second, the speedup in general increases as the cluster size grows.
The main reason is that network is the bottleneck for PS even with 10Gb bandwidth, and adding workers to accelerate computation only slightly reduces the total training time. On the contrary, \sys cuts the communication time significantly by replacing gradients with activations, and thus the benefit of parallelism from adding workers prevails.

Figure~\ref{fig:1ps_fc} and \ref{fig:1ps_max} corroborate the above analysis. 
We only show the results for AlexNet and VGG-16 for brevity.
Figure~\ref{fig:1ps_fc} shows that PS requires $\sim$100x larger data transfer than \sys for FC layers, which is the main reason for \sys's training time improvement. 
FC-Data increases with more workers because in one {\em iteration}, the amount of FC layer data transfer depends on the total batch size and scales with the number of workers.
Figure~\ref{fig:1ps_max} shows the Total-Data comparison. 
Observe that PS's total data transfer per epoch is over 4x larger than \sys's for all cluster sizes (the figure is in log-scale). In other words, PS needs to perform data transfer that is at least 4 times larger than \sys. 
Total-Data of PS does not change with number of workers because although FC-Data per iteration increases with more workers, the number of iterations decreases proportionally. Essentially Total-Data in PS only depends on the total number of samples.
Total-Data of \sys increases slightly with more CONV workers because the CONV worker communication takes more rounds to complete based on the analysis in \cref{sec:perf_model}. 
% The two figures also show that \sys's saving in data transfer becomes larger as the cluster grows, which in turn also attributes to the improving speedup in training time. 

\noindent{\bf Quick recap:}
We demonstrate that in production settings with 10G bandwidth and state-of-the-art data center GPU, \sys provides significant improvements in training time over PS, which increases as the cluster size grows.

%% file: numerical.tex
%!TEX root = main.tex
\section{Numerical Results}
\label{sec:simulation}

% Lastly, based on the experimental data and performance model derived in \cref{sec:perf_model}, we show numerically that even with abundant bandwidth at 40G and 100G, \sys is able to provide \xr{more than 2x} \hx{xxx} speedup.
\begin{comment}
\xr{Generally, some research institutes and industrial institutes deploy their missions on cloud providers, which is easy to scale the cluster of training. We survey the top-2 cloud providers in the world, AWS and Azure. The table~\ref{table:bandwidth} shows bandwidth resources of the common VMs in deep learning. Because of the distributed training, we show the VMs just with one GPU card. We find the bandwidth of p3.2xlarge is up to 10Gbps which is the burst bandwidth. In our evaluation part, we adopt the NC6 and p3.xlarge, which presents the typical VMs in the distributed training.}

\begin{table}[]
\begin{tabular}{|c|c|c|c|}
\hline
Instances      & GPU      & Ethernet       & Provider \\ \hline
Standard NC6   & K-80      & $\sim$2.6Gbps  & Azure    \\ \hline
% Standard NC24r & k80*4    & $\sim$4.6Gbps  & Azure    \\ \hline
P2.xlarge      & K-80      & $\sim$1.24Gbps & AWS      \\ \hline
p3.2xlarge     & V-100     & $\sim$10Gbps   & AWS      \\ \hline
% p3.16xlarge    & V100 * 8 & $\sim$25Gbps   & AWS      \\ \hline
\end{tabular}
\caption{\small Combinations of models and datasets used in the evaluation. Batch size is for each worker. }
% \vspace{-5mm}
\label{table:bandwidth}
\end{table}
\end{comment}
Our evaluation so far covers small and medium clusters with typical GPU and bandwidth settings in public clouds. 
In reality, companies may use larger clusters for training with huge data. 
Also some clusters may deploy faster networks (40GbE, 100GbE, or RDMA) \cite{Mellanox}.  
Yet it is difficult for us to access these hardware resources especially in a large scale. 

Therefore in this section, we numerically evaluate \sys in large-scale settings based on the empirical data from testbed experiments in \cref{sec:eval} and the performance model in \cref{sec:perf_model}. 
We first develop a performance model for PS systems, similar to the one for \sys, to estimate the training time and throughput. 
Then we verify the fidelity of the two models with experiments in public clouds. 
Finally, we estimate the throughput of both systems in large-scale settings using the performance models.

\subsection{Performance Models and Verification}
\label{sec:verification}
% We validate \sys in two real cloud providers, Azure and AWS. For the standard\_NC6 instance from Azure, the bandwidth among instances is about 2.6GbE. The p3.2xlarge in the AWS provides the 10GbE Ethernet. Both two kind bandwidth presents the popular network settings in the public cloud providers. Although we get the obvious benefits for our experiments, we want to show that \sys can still performance well in the larger bandwidth. We provide the performance model to simulate the training with 20GbE and 40GbE Ethernet. In the simulation, we still use the model we mentioned before.
The performance model for \sys is explained in detail in \cref{sec:perf_model}. 
Here we develop a similar model for PS. 
Denote the computation time to process a batch of samples on one worker, including the forward pass and backpropagation, as $T_{ps}$.
With $n_s$ servers, each is responsible for an equal share of ${P}/{n_s}$ parameters. 
The communication time in each iteration composes of (1) each server receiving ${P}/{n_s}$ gradients from each worker, which takes $n_w {P}/{n_s}B$ time, and then (2) each server sending ${P}/{n_s}$ parameters to each worker, which takes $n_w {P}/{n_s}B$ as well. Thus, training time per iteration is
\begin{equation}
	t_{ps} = \frac {2n_wP}{n_sB} + T_{ps}.
\label{eqn:PST}
\end{equation}
Training throughput is then
\begin{equation}
throughput_{ps}(n_w, n_s) = \frac {n_w K}{t_{ps}}.
\label{eqn:TH}
\end{equation}

\begin{figure}[t]
% \vspace{-4mm}
\centering
\subfloat[Stanza]{\label{}{\includegraphics[width=0.5\linewidth]
{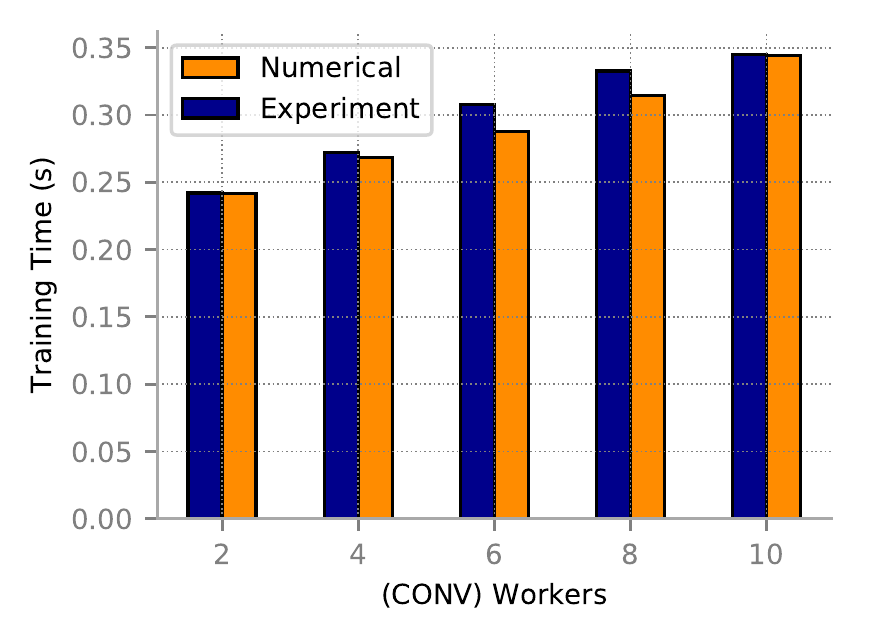}}}\hfill
\subfloat[Parameter Server]{\label{}{\includegraphics[width=0.5\linewidth]
{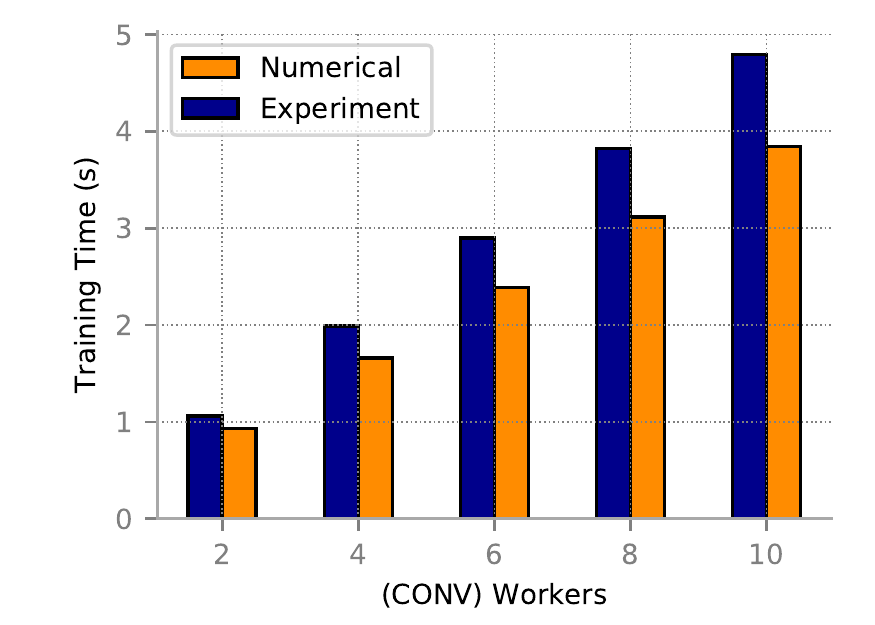}}}\hfill
% \subfloat[Max Comm Requirement]{\label{MXCR}{\includegraphics[width=0.33\textwidth]{2PS_MCR}}}
\caption{\small Accuracy of the performance models using AlexNet on ImageNet-12 in the EC2 cluster. }
\label{fig:veri}
% \vspace{-4mm}
\end{figure}

We measure $T_{ps}$ in PS and the single worker computation times $T_c$ and $T_f$ for \sys in our EC2 GPU cluster, and feed them to the performance models to estimate the per-iteration training time. 
The model estimation is compared against the empirical data to verify their fidelity.
We use AlexNet on ImageNet-12 with the same setup in \cref{sec:setup} here. 
Figure~\ref{fig:veri} shows the results with varying number of (CONV) workers. 
Observe that our models are fairly accurate. 
Comparing to experimental results, our model in \eqref{eqn:MT} estimates the per-iteration training time of \sys with $\sim$5\% errors, and Equation~\eqref{eqn:PST} estimates PS training time with $\sim$20\% errors. 
The errors may be attributed to an array of factors our models do not capture, such as fluctuating network bandwidth, memory copy time between CPU and GPU, etc.
More importantly, they are acceptable for our purpose here because our model underestimates PS's training time, and thus serves as a conservative lower bound for the speedup comparison against \sys.

\begin{figure*}[th]
\centering
\subfloat[AlexNet]{\label{fig:alex_simu}{\includegraphics[width=0.33\textwidth]{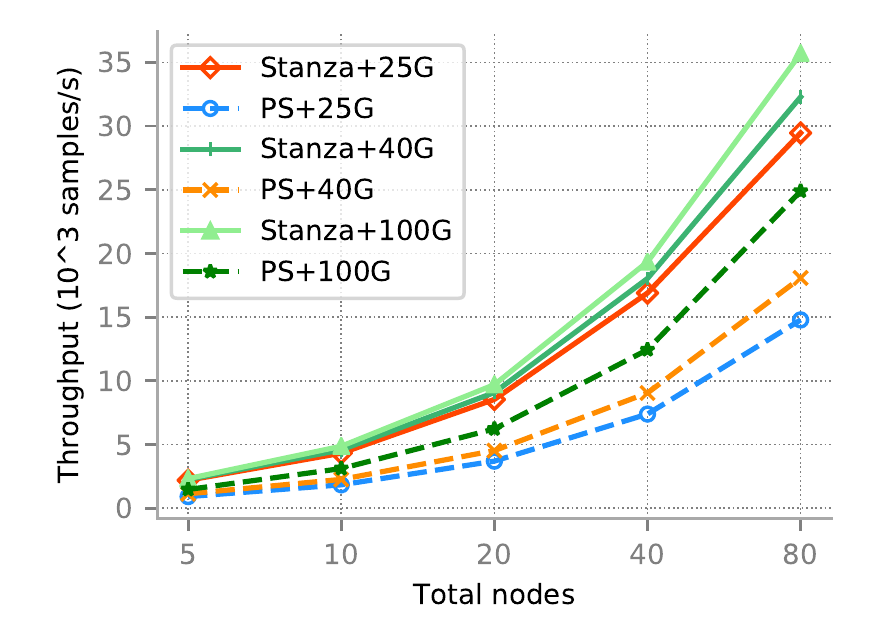}}}\hfill
\subfloat[VGG-16]{\label{fig:vgg16_simu}{\includegraphics[width=0.33\textwidth]{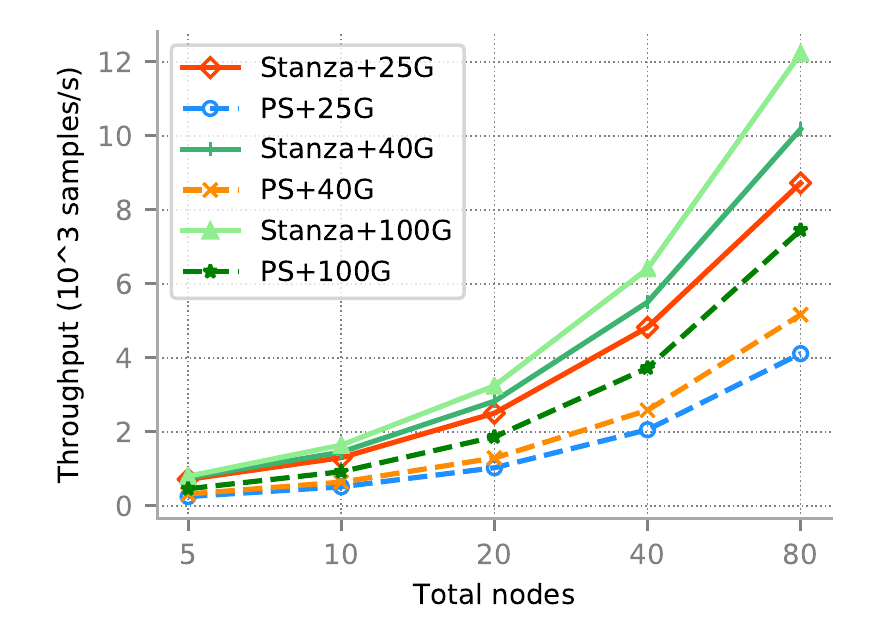}}}\hfill
\subfloat[VGG-19]{\label{fig:vgg19_simu}{\includegraphics[width=0.33\textwidth]{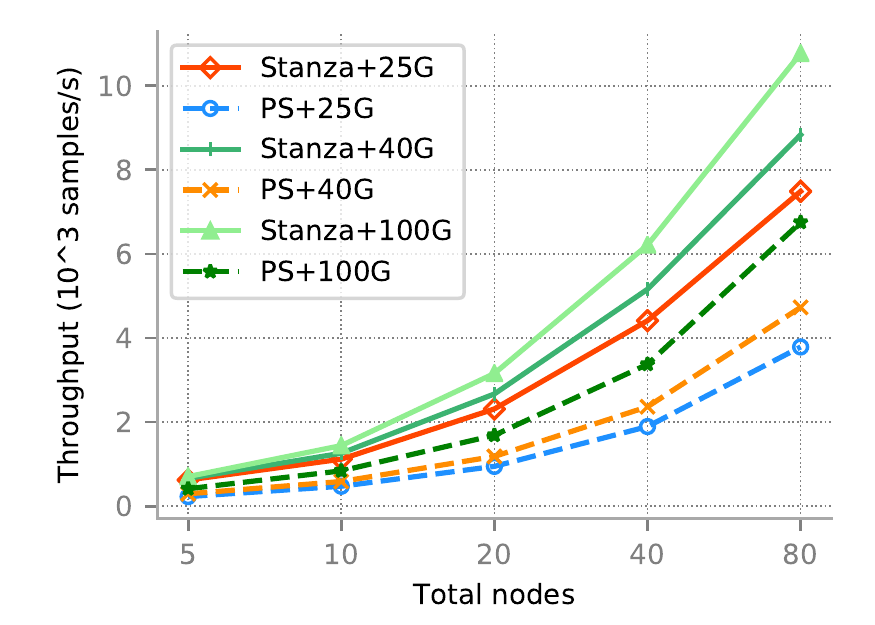}}}
\vspace{-3mm}
\caption{\small Numerical results for different CNNs in large-scale clusters using the performance models. \sys and PS use 5, 10, 20, 40, and 80 nodes in total. The y-axis shows throughput in $10^3$ samples per second. The x-axis is not in linear scale. }
\label{fig:simulation}
\end{figure*}

\subsection{Numerical Simulation}
\label{sec:sim_results}

We now proceed to estimate the throughput performance of \sys and PS in large-scale clusters.
We use AlexNet, VGG-16, and VGG-19 on ImageNet-12, and obtain empirical data from our EC2 experiments for the model constants $T_{ps}$, $T_{c}$ and $T_{f}$. 
They reflect the computation performance on ImageNet-12 with Tesla V100 GPU, the fastest data center GPU presently.   
We assume each node has a single GPU. 
We vary the network bandwidth from 25G to 40G and 100G.

\sys and PS use the same number of nodes in total, which varies from 5 to 80. 
\sys applies the node assignment algorithm in \cref{sec:node_assignment} to
determine the optimal number of FC workers in each setting, which can be
larger than one especially for large clusters.\footnote{For CONV workers, we
adopt the batchsizes used in the original papers of the models. Because a
large batchsize may cause GPU memory overflow, we also consider the capacity
of GPU memory in node assignment.} Similarly PS also uses the model in
\cref{sec:verification} to obtain the throughput-maximizing node assignment in
each case. For this reason the two systems may have different total batch
sizes, and thus we focus on throughput instead of training time as the
performance metric.

\begin{comment}
The number of FC workers are 1, 1, 2, 4, 4 separately. 
The ratio of workers and servers in PS is 4:1, i.e. from 1 to 16.
\end{comment}
\begin{comment}
Figure~\ref{fig:simulation} shows the numerical results.
We make several observations.
First, \sys still outperforms PS significantly in a large cluster with fast network. With 80 nodes and 100G bandwidth for AlexNet, the throughput improvement is $\sim$1.4x.
Second, as the network bandwidth increase, communication becomes less of a problem, both systems have the performance improvement. At 25G bandwidth, training AlexNet, VGG-16, and VGG-19 can be improved by $\sim$2.29x, $\sim$2.35x, and $\sim$2.33x, respectively using \sys with 40 nodes.
\end{comment}  

Figure~\ref{fig:simulation} shows the numerical results. 
We make several observations.
First, \sys still outperforms PS significantly in a large cluster with fast network. With 80 nodes and 40G bandwidth for example, the throughput improvement is $\sim$1.79x for AlexNet, $\sim$1.97x for VGG-16, and $\sim$1.87x for VGG-19.
Second, as the network bandwidth increases, communication becomes less of a problem, and \sys's improvement decays. 
At 25G bandwidth, training AlexNet, VGG-16, and VGG-19 can be improved by $\sim$2.29x, $\sim$2.35x, and $\sim$2.33x, respectively using \sys with 40 nodes.
When the bandwidth is 100G, \sys's benefit decreases to $\sim$1.55x, $\sim$1.72x, and $\sim$1.84x, respectively for the three CNNs. 

Third, \sys enjoys better scalability compared to PS. 
With 100G bandwidth, \sys delivers 2x throughput improvement for AlexNet (resp. VGG-19) when the number of nodes increases from 40 to 80; PS only achieves 1.85x (resp. 1.73x) improvement in the same scenario. 
% For VGG-19, with 100G bandwidth \sys delivers 2x throughput improvement while PS only achieves 1.73x from 40 nodes to 80.
Lastly, a related observation is that PS's scalability largely depends on the network bandwidth.
Without enough bandwidth PS scales poorly due to again the communication bottleneck. 
For example with 25G bandwidth, PS has 1.73x, 1.81x, and 1.70x throughput benefits from 40 nodes to 80 for AlexNet, VGG-16, and VGG-19, respectively.
With 100G bandwidth, the gains are better at 1.85x, 1.91x, and 1.73x, respectively. \sys on the other hand consistently delivers 2x throughput scalability even with 25G bandwidth for all three CNNs.

\noindent{\bf Quick recap:}
Our numerical study shows that \sys achieves $\sim$2x better throughput over PS in large-scale clusters with up to 100G bandwidth. \sys also has better scalability due to its ability to remove the communication bottleneck.

%% file: dis.tex
%!TEX root = main.tex
\section{Discussion}
\label{sec:dis}

% We now discuss several directions along which \sys can be extended to further improve its performance and practicality. 
We discuss several concerns one may have about \sys.

\noindent{\bf Beyond Parameter Server.}
Recently, to cope with the scalability issue of PS systems, new 
communication strategies for distributed training have been proposed and
deployed in some
cases.
% traditional high performance computing (HPC) communication strategies attract more people's attention in distributed deep learning. 
Uber proposes Horovod \cite{sergeev2018horovod} which uses ring-reduce
communication. Baidu's internal system uses optimized allreduce communication 
\cite{baidu_allreduce}. This is orthogonal to our approach of
reducing data transfer by layer separation, because both systems still rely on
data parallelism with each node going through the complete model. For example
we can use Horovod on the CONV workers to accelerate their communication.
\sys also applies to these non-PS systems to further improve training time,
though we leave it as future work the implementation and evaluation of such an
extension.
\begin{comment}
\sys also adopts the allreduce communication strategy for CONV workers communication. Both ring-reduce and allreduce are from the Message Passing Interface (MPI).
There are two reasons can explain why \sys doesn't completely use allreduce among all nodes.
First, the challenge for the MPI is it can not work well in the asynchronous environment, for example, both Horovod and Baidu's system are running in synchronous environment. In the contrast, it is easy to convert \sys from synchronous environment to asynchronous environment. \sys just needs one more node to be a parameter server which only asynchronously communicates with any Conv node. We give the details in section \cref{sec:conclusion}.
Second, the nature of reduce algorithm is to reduce the data through merging the data with same index among different nodes, while the activation among different nodes cannot be merged, so the characteristic determines that \sys cannot use allreduce between FC worker and CONV worker.
\end{comment}

\noindent{\bf Beyond 10G Network Bandwidth.}
High bandwidth networks at 40G or 100G are deployed in some private data
centers to speed up distributed training. However, 10Gbps network is the
mainstream in public clouds and most private DL clusters. Clearly \sys's gain
is less substantial with higher bandwidth since the impact of data transfer is
smaller. Yet as we have numerically shown in \cref{sec:simulation}, \sys still
provides over 55\% gain over the current PS systems with 100G bandwidth. As
GPUs are improving at a rapid pace, we believe \sys is instrumental for many
deployment scenarios.
% We observe that V100 GPU is available in the most large public cloud providers, like AWS and Azure. Meanwhile, the capacity of network for the instance equipped with V100 is up to 10Gbps. With the infinite bandwidth network popularizing in the public cloud provider, the faster computing device will also appear. So, \sys can have a considerable performance during the follow-up session.
% We also observe that the 40Gbps and 100Gbps network are starting to be deployed in the industry. However, the 10Gbps network is still widely used in the most companies. From the \cref{sec:simulation}, we can find that \sys is still performance faster training than PS architecture even with 100Gbps bandwidth. In the simulation, we adopt the 40Gbps and 100Gbps network which are not sold on the public cloud provider, meanwhile the V100 GPU the fastest GPU is already sold on the public cloud provider. When the 40Gbps and 100Gbps network are widely used in the public cloud provider, the faster computing device will also appear.

\noindent{\bf Beyond CNNs.}
We have focused on CNNs in this paper. The idea of layer separation works for
many other DL models, where nodes exchange only the activations of the
boundary layer instead of the full set of model parameters.
% so long as the layers exhibit different computation and
% communication characteristics. For example, recurrent neural network (RNN) is
% also composed of fully connected layers, long short term memory (LSTM) layers,
% convolutional layers and so on \cite{hannun2014deep,press2016using}. We aim to
% explore the use of \sys in general DL models more thoroughly.
For example, recurrent neural networks (RNNs)
\cite{hannun2014deep,press2016using} have recently received much attention
with many applications
\cite{mikolov2010recurrent,graves2014towards,liu2014recursive}. Generally, an
RNN model consists of cells each with the same multilayer perceptron (MLP)
models, which have multiple FC layers without CONV layers. To show that our
idea is also beneficial for RNNs, we consider the MLP in the RNN cells without
dependency. We conduct a simple experiment, where an MLP is separated after
the first 2 hidden layers, and we use just one node to train the {last hidden
layer and the output layer} while the rest of nodes to train the first two
hidden layers. As Table~\ref{tb:MLP} shows, layer separation achieves 3.1x to
4.2x speedups over PS under the same resource and hyperparameter settings
without sacrificing the accuracy. We plan to investigate in future work how to
deal with the dependency between RNN cells in order to fully extend \sys to
RNN and other DL models.

\begin{table}[h]
\centering
\footnotesize
\resizebox{\columnwidth}{!}{
\begin{tabular}{ccccccl}
\hline
\multirow{2}{*}{Nodes} & Base                  & Total                 & 
\multirow{2}{*}{System} & Training       & \multicolumn{1}{l}{\multirow{2}{*}{Speedup}} & \multirow{2}{*}{Accuracy} \\
                       & LR                    & batchsize             &                         & time (s/epoch) & \multicolumn{1}{l}{}                          &                           \\ \hline
\multirow{2}{*}{2}     & \multirow{2}{*}{0.02} & \multirow{2}{*}{256}  & LS &
16.8 & \multirow{2}{*}{3.1}                          & 53.37\%                   \\
                       &                       &                       & PS                      & 52.2           &                                               & 52.04\%                   \\ \hline
\multirow{2}{*}{4}     & \multirow{2}{*}{0.04} & \multirow{2}{*}{512}  & LS    
& 14.7           & \multirow{2}{*}{3.7}                          & 49.83\%                   \\
                       &                       &                       & PS                      & 54.6           &                                               & 48.97\%                   \\ \hline
\multirow{2}{*}{8}     & \multirow{2}{*}{0.08} & \multirow{2}{*}{1024} & LS    
& 12.7           & \multirow{2}{*}{4.2}                          & 44.49\%                   \\
                       &                       &                       & PS                      & 53.0           &                                               & 45.79\%                   \\ \hline
\end{tabular}}
% \caption{}
\caption{Statistics for training MLP with layer separation (LS in the table) and
PS. We use an MLP with 3 {hidden} layers each with 1024, 1024, and 4096 hidden
units. We separate the model after the second {hidden} layer. We train the
last hidden layer and the output layer with only one node, and the first two
{hidden} layers with the rest of nodes. We use vanilla SGD to train the MLP
for 100 epochs with learning rate decaying every 30 epochs. We use the Azure
testbed as described in \cref{sec:setup} and CIFAR-10 as the dataset. }
\label{tb:MLP}
\vspace{-4mm}
\end{table}

\begin{comment}
\noindent{\bf \xr{response 6.1. Cost for Searching Deployment Policy}}
Training a complex model is expensive, it can spend weeks or days time. For example, Google released Bert\cite{devlin2018bert}, a language representation model, which takes weeks to train the model on the TPU cluster. So, it is urgent to speed the training period. Comparing with the long training period, it only takes little time to pre-tuning the deployment policy for the training in \sys. There are two part cost. First, the \sys may need to sample the training time for the separated model. This cost depends on how to separate the model, which is flexible, but it is not too long. In our experiment, we find that the ration of parameters to activations is the index. The index is larger, the more communication cost is saving. The second part is to compute the optimal deployment, which is a result from equation \ref{eqn:MT} and stable. The pre-tuning phase is short, which doesn't have on really impact on the whole training cost.
\end{comment}

%% file: related.tex
%!TEX root = main.tex

\section{Related Work}
\label{sec:related}

We survey related work in this section.

\noindent\Emph{Parameter Server Architecture.}
Parameter server has attracted much attention in both academia and industry
since Distbelief \cite{dean2012large}. Li et al. \cite{li2014scaling} propose
a general PS system design that supports flexible model consistency, elastic
scalability, and fault tolerance. Its variants and extensions have since been
widely deployed. Examples include popular DL frameworks such as Tensorflow
\cite{abadi2016tensorflow} and MXNet \cite{chen2015mxnet}.

Several approaches have been proposed to deal with the communication problem
in {distributed training}. We discuss some important ones other than those mentioned in
\cref{sec:dis} now.

\noindent\Emph{Communication Strategies.}
% Training time can be improved by designing better communication strategies. 
Xie et al. \cite{XKZH16} propose sufficient factor broadcasting that transmits
vectors to fully re-construct the parameter matrix. The amount of data
transfer thus scales linearly instead of quadratically with the dimensions of
the parameter matrix. Poseidon \cite{203269} exploits layered structures of DL
models to overlap communication and computation and hide the communication
cost. {Both strategies can be applied in \sys to further reduce the
communication bottleneck.}

Another approach is to use decentralized SGD algorithm \cite{lian2017can}
instead of centralized SGD. Nodes can be arranged in a ring topology. Each
node only communicates with its neighbors instead of the entire cluster to
exchange gradients and parameters at each iteration. This approach certainly
speeds up the training time per iteration. Yet it demands multiple iterations
to populate the parameters across the entire cluster, and the convergence
speed may degrade in the end in practice.

\noindent\Emph{Synchronization.}
Current PS systems generally use Bulk Synchronous Parallel (BSP)
\cite{valiant1990bridging} to ensure model consistency among workers. BSP
tends to prolong training time when stragglers are present. Asynchronous
Parallel (ASP) and Stale Synchronous Parallel (SSP)
\cite{ho2013more,wei2015managed} alleviates the impact of
stragglers by removing the barriers in BSP. Yet they
deteriorate the convergence speed and model performance due to the staled
parameter information \cite{chen2016revisiting}. How to achieve a good
tradeoff between convergence and synchronization is still an open problem.
\sys adopts BSP in order to achieve fast convergence and better performance. 
It can also be readily extended to use ASP or SSP.

\noindent\Emph{Gradient Compression.}
Another interesting approach is to compress the gradients and parameters. 
% via sparsification and quantization. 
Gradient sparsification exploits the fact that the parameter matrix is sparse
and some weights are small. Aji et al. \cite{aji2017sparse} propose to
truncate the insignificant gradients and only transmit the larger ones. Hsieh
et al. \cite{hsieh2017gaia} use a similar method that only sends important
gradients to servers in a WAN setting. Gradient quantization reduces the
number of bits to represent gradients and parameters \cite{LHMW18}. CNTK
\cite{yu2015microsoft} reduces the size of gradients via 1-bit quantization,
which performs fairly well in speech recognition \cite{seide20141}. TernGrad
\cite{wen2017terngrad} uses 2-bit quantization and is shown to have little
accuracy loss.
\sys is orthogonal to these efforts, and can use them to further alleviate the
communication cost.

%% file: conclusion.tex
%!TEX root = main.tex
\section{Conclusion and Future Work}
\label{sec:conclusion}
We have presented the design and implementation of \sys, a
communication-efficient DL system for distributed training.
\sys utilizes the unique characteristics of CNNs to separate the training of
CONV and FC layers. FC layer parameter exchange is limited to among a few FC
workers as a result, and the bulky data transfer across the network in
conventional parameter server systems is largely removed. Testbed experiments
in Azure and EC2 show that \sys improves training time substantially over
parameter server systems with latest datacenter GPUs and 10G bandwidth.
Numerical studies also indicate that \sys still provides modest speedup even
with 100G bandwidth in large-scale clusters.

In this work we decouple the DL model at the boundary of CONV and FC layers,
which is intuitive and easy to implement. As future work, other strategies of
model decomposition \cite{coates2013deep} can be considered to further
re-distribute computation. For example multiple decompositions can be realized
by using additional machine groups to train the specific model partitions. The
performance model may also be modified correspondingly for optimal node
assignment. Other than layer based model decomposition, it is also possible to
use more fine-grained model decomposition. For example it has been shown that
scheduling the different operations of DL training to different computing
devices can also improve training speed, though finding the optimal device
placement itself is computationally expensive \cite{mirhoseini2017device}.
Extending \sys to non-PS systems and general DL models is also a promising
direction of future work as detailed in \cref{sec:dis}.

% In this sense our layer based decomposition may be a more efficient and
% promising solution.

\begin{comment}
\xr{
Furthermore, we use BSP to ensure model consistency among workers in \sys. 
However, the distributed training has the small probability to face the straggler which will dwarf the training speed. As a future work, we want to implement ASP in the \sys, which can reduce the impact brought by the straggler. What's more, ASP can further reduce communication pressure in \sys.
It is convenient to convert \sys from BSP to ASP. We should add one more node in \sys. This node is responsible for asynchronously updating the parameters in CONV workers. In ASP pattern, there is no inter communication among CONV workers. Each worker should send the activation to the FC worker and update parameters with the new added node.
} 
\end{comment}